\documentclass[10pt,twocolumn]{article} 
\usepackage{preprint}
\colortheme{green} 

\usepackage{multirow}
\usepackage{subcaption}
\usepackage{titletoc}
\usepackage{makecell}

\runningtitle{Mumcu et al.}

\definecolor{titlegreen}{HTML}{15803D}

\definecolor{letterL}{HTML}{4C9E2E}  
\definecolor{letterA}{HTML}{9E772E}  
\definecolor{letterO}{HTML}{802E9E}  
\definecolor{letterD}{HTML}{2E649E}  

\newcommand{\LAOD}{\textbf{\textcolor{letterL}{L}\textcolor{letterA}{A}\textcolor{letterO}{O}\textcolor{letterD}{D}}}

\newcommand{\titlerest}[1]{\textcolor{mutedtext}{\fontsize{15pt}{20pt}\selectfont #1}}

\ptitle{%
  \titlerest{M}\textcolor{letterL}{L}\titlerest{LM-Guided} \textcolor{letterA}{A}\titlerest{utonomous} \textcolor{letterO}{O}\titlerest{bject} \textcolor{letterD}{D}\titlerest{etection for}
  \titlerest{Open-World Understanding}%
}

\definecolor{titlegreen}{HTML}{15803D}
\definecolor{modernred}{HTML}{B91C1C}

\authormarkerstyle{tag}
\affiliationcolor{USF}{titlegreen}
\affiliationcolor{MERL}{modernred}

\pauthor{Furkan Mumcu}{USF}{furkan@usf.edu}
\pauthor{Michael J. Jones}{MERL}{}
\pauthor{Anoop Cherian}{MERL}{}
\pauthor{Yasin Yilmaz}{USF}{}
\paffiliations{%
  \textsuperscript{\textcolor{titlegreen}{\scriptsize\sffamily\bfseries\upshape USF}}\,Department of Electrical Engineering, University of South Florida
  \\
  \textsuperscript{\textcolor{modernred}{\scriptsize\sffamily\bfseries\upshape MERL}}\,Mitsubishi Electric Research Laboratories%
}

\pabstract{%
Object detection typically relies on predefined category vocabularies, while
open-world and open-vocabulary approaches remain limited by generic unknown
labels or externally specified detection concepts. We propose \LAOD{}, an
M\textcolor{letterL}{L}LM-Guided
\textcolor{letterA}{A}utonomous
\textcolor{letterO}{O}bject
\textcolor{letterD}{D}etection framework in which a Multimodal Large Language
Model (MLLM) generates an image-specific detection vocabulary that is grounded
by an open-vocabulary detector. This decomposition separates semantic concept
discovery from spatial localization and removes the need to manually enumerate
target categories at inference time. We evaluate \LAOD{} using Class-Agnostic
Average Precision (CAAP), Semantic Naming Average Precision (SNAP), and
Semantic Localization Average Precision (SemLoc-AP), which measure
localization, semantic naming, and their correct association, respectively.
Experiments on COCO-Val, LVIS-minival, and COCO-OOD across multiple MLLMs and
detectors demonstrate strong performance across common, long-tailed, and
out-of-distribution objects, while showing that grounding quality remains a
major determinant of end-to-end performance. \LAOD{} also substantially outperforms both using MLLMs directly for object
detection and dedicated models trained specifically for open-ended detection. Overall, the results
support a modular formulation of autonomous object detection in which MLLMs
determine what to detect and specialized open-vocabulary detectors determine
where those concepts occur.

\par\vspace{0.6em}
Project Page: \url{https://github.com/furkanmumcu/LAOD}

}

\begin{document}

\makepreprinttitle

\section{Introduction}

Object detection is a fundamental computer vision task that aims to localize and classify objects within an image. Traditional object detectors, however, typically operate under a closed-world assumption: the set of object categories is fixed during training~\cite{ren2015faster,redmon2016you}. Adapting such systems to new categories or changing detection requirements therefore often requires additional training or fine-tuning, limiting their flexibility in open and dynamic environments.

Two major paradigms have emerged to relax this assumption: Open-World Object Detection (OWOD) and Open-Vocabulary Object Detection (OVOD). OWOD extends conventional detection by identifying objects that do not belong to the known training categories~\cite{joseph2021open,owdetr,zohar2023prob,ma2023cat,sun2024orthogonal}. However, these novel objects are typically represented under a generic ``unknown'' category rather than assigned meaningful semantic names~\cite{joseph2021open,zohar2023prob,yavuz2024o1o}. Moreover, open-world detection introduces an inherent ambiguity in determining what should constitute a detection target. For example, given an image containing a person and a car, a detector may be expected to identify only the two primary objects, or also finer-grained entities such as a shirt, head, tire, or door. Defining such granularity exhaustively through fixed annotations is difficult in unconstrained environments.

OVOD addresses a complementary limitation by allowing detectors to localize arbitrary categories specified through natural-language queries~\cite{gu2021open,zhou2022detic,zhong2022regionclip,du2022detpro,li2022glip,kuo2023fvlm}. This greatly expands the range of concepts that can be detected without retraining~\cite{wu2023cora,wu2023baron,yao2023detclipv2,shi2023edadet}, but the detection vocabulary must still be supplied externally in standard OVOD formulations~\cite{minderer2022owlvit,owlv2,gdino,yoloworld,cho2024decola}. In other words, an OVOD model can localize a category once it is asked to search for it, but it does not generally determine by itself which categories are relevant to the current scene. This dependence on an externally specified vocabulary limits autonomous discovery when the objects of interest are not known in advance.

\begin{figure}[t]
  \centering
  \includegraphics[width=0.9\linewidth]{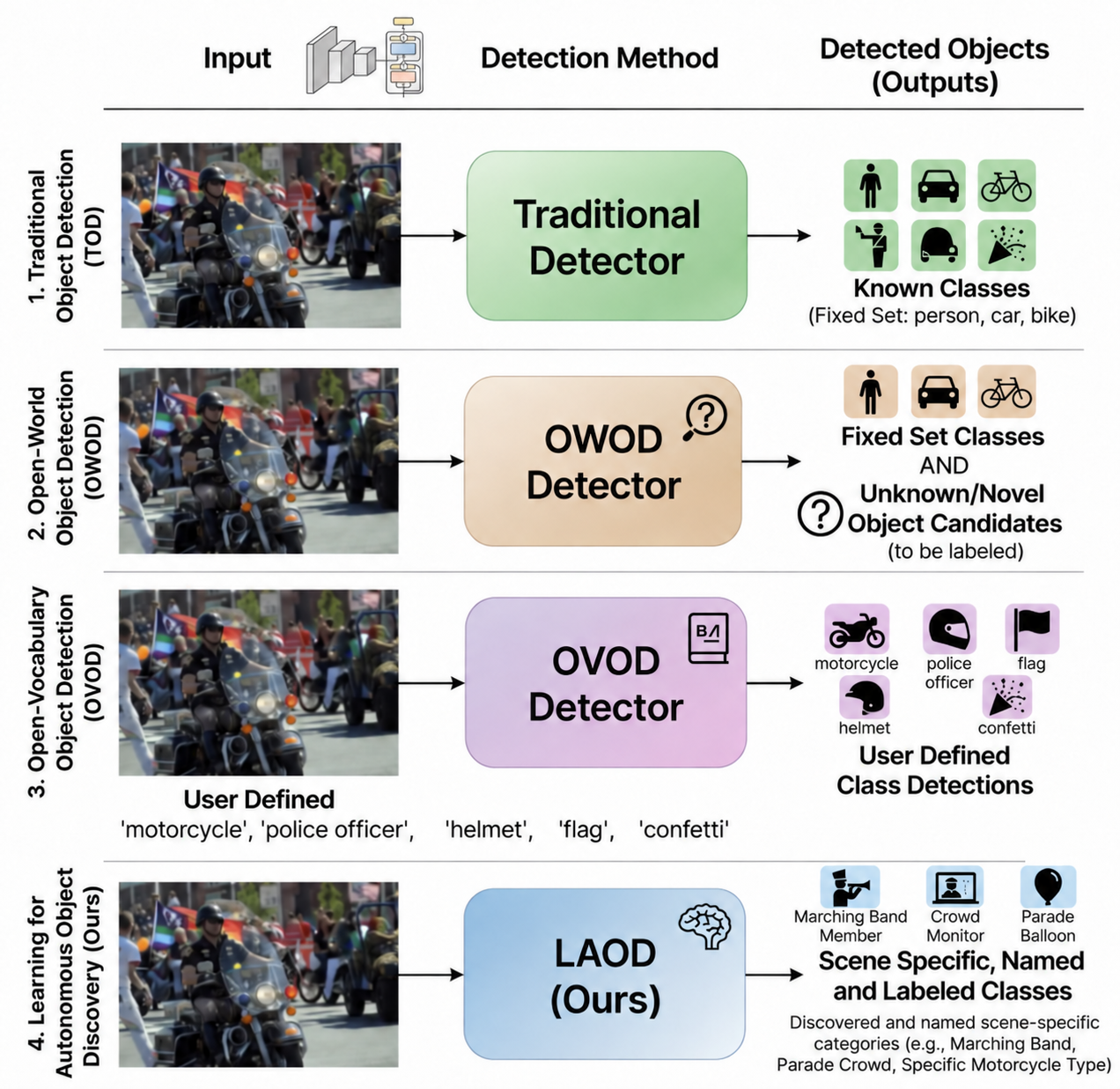}
  \caption{Comparison of object detection paradigms: traditional object detection, OWOD, OVOD, and our proposed autonomous \LAOD{} framework.}
  \label{fig:comp}
\end{figure}

This decomposition separates two complementary capabilities. The MLLM
determines \emph{what} concepts are relevant to the scene, while the
open-vocabulary detector determines \emph{where} those concepts occur. Although
modern MLLMs can also generate bounding-box coordinates directly
~\cite{bai2023qwenvl,peng2024kosmos2,chen2023shikra,you2024ferret,
rasheed2024glamm,zhang2024llavagrounding,wang2024qwen2vl,deitke2025molmo},
our decomposition does not assume that spatial prediction is beyond their
capabilities. Rather, it allows semantic scene interpretation and vocabulary
construction to be handled by the MLLM while dense instance enumeration,
precise localization, and confidence-ranked prediction are delegated to a
specialized detector. We explicitly evaluate whether this separation is
beneficial by comparing \LAOD{} with the same MLLMs used directly as object
detectors.

The resulting system requires no predefined category set at inference time and
can adapt its detection vocabulary to each image. At the same time, the prompt
supplied to the MLLM provides a natural mechanism for controlling the desired
detection scope. A general prompt can encourage broad scene understanding,
whereas a more specific prompt can restrict detection to particular types of
objects or levels of granularity. Thus, autonomy in \LAOD{} refers specifically
to autonomous construction of the detection vocabulary rather than the absence
of task-level guidance.

This formulation also raises an evaluation challenge. Conventional detection
metrics assume a fixed correspondence between predicted and ground-truth class
labels~\cite{uap,coco}, whereas \LAOD{} is free to generate semantically
appropriate names that may differ lexically from benchmark annotations. We
therefore evaluate three complementary aspects of the resulting detections.
Class-Agnostic Average Precision (CAAP) measures whether objects are
successfully localized independently of their predicted names, while Semantic
Naming Average Precision (SNAP) measures semantic agreement between generated
and ground-truth labels independently of spatial localization. These two views,
however, do not determine whether an appropriate semantic name is assigned to
the particular object that was localized. We therefore additionally introduce
Semantic Localization Average Precision (SemLoc-AP), which requires the same
ground-truth instance to satisfy both spatial and semantic matching criteria.
Together, these metrics distinguish localization quality, semantic naming
quality, and their joint correctness.

In summary, this work makes the following contributions:

\begin{itemize}
    \item We introduce \textbf{M\textcolor{letterL}{L}LM-Guided \textcolor{letterA}{A}utonomous \textcolor{letterO}{O}bject \textcolor{letterD}{D}etection (\LAOD{})}, a modular framework in which an MLLM autonomously generates an image-specific detection vocabulary that is subsequently grounded by an open-vocabulary object detector, removing the need for a predefined or manually enumerated category set at inference time.

    \item We systematically study the two components of this framework across multiple MLLMs and open-vocabulary detectors on COCO, LVIS, and COCO-OOD, demonstrating that \LAOD{} is not tied to a particular MLLM or detector and analyzing how each component affects performance. We further compare against using MLLMs directly for object detection, showing the benefit of separating semantic concept discovery from spatial grounding.

    \item We investigate the controllability and prompt sensitivity of autonomous detection through language instructions. The detection scope and granularity can be modified through task-level prompting without retraining the detector, while experiments show that generic prompt variations have limited effect and benchmark-specific prompt specialization does not transfer consistently across datasets.

    \item We evaluate \LAOD{} using CAAP for class-agnostic localization and introduce SNAP and SemLoc-AP to evaluate semantic naming and joint semantic-localization correctness when the predicted vocabulary is generated dynamically rather than selected from a fixed set of benchmark classes.
\end{itemize}
\section{Related Work}

\begin{figure*}[t]
  \centering
   \includegraphics[width=0.8\linewidth]{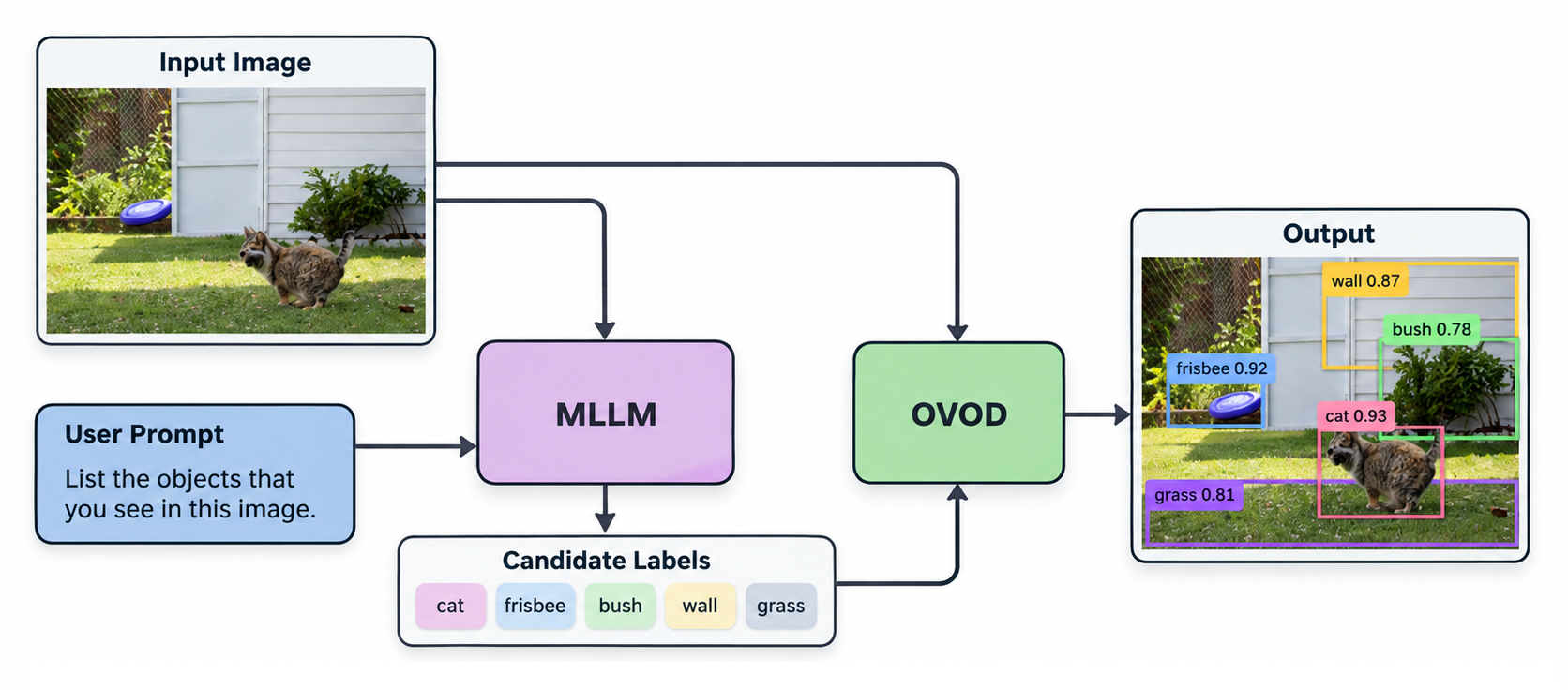}
\caption{Overview of the M\textcolor{letterL}{L}LM-Guided \textcolor{letterA}{A}utonomous \textcolor{letterO}{O}bject \textcolor{letterD}{D}etection (\LAOD{}) framework. Given an input image, the MLLM generates an image-specific detection vocabulary, which is then grounded by an open-vocabulary object detector to localize the corresponding objects.}
\label{fig:method}
\end{figure*}

Object detection has traditionally been formulated under a closed-world assumption, where models are trained to recognize a fixed set of predefined object categories \cite{ren2015faster,redmon2016you}. Although highly effective within their training distributions, such detectors generally require additional training or fine-tuning to incorporate new categories or adapt to substantially different detection requirements. This limitation has motivated detection paradigms that relax the dependence on a fixed vocabulary.

\textbf{Open-World Object Detection.}
Open-World Object Detection (OWOD) extends conventional detection by requiring models to identify both known objects and objects belonging to previously unseen categories
\cite{joseph2021open,owdetr,zohar2023prob,ma2023cat,sun2024orthogonal}.
Unknown objects are generally localized and assigned a generic ``unknown'' label rather than a specific semantic category
\cite{joseph2021open,zohar2023prob}.
Related Unknown Object Detection (UOD) methods focus more directly on separating unknown objects from known categories and background regions
\cite{unsniffer,undetr,yavuz2024o1o}.
These approaches provide an important capability for discovering objects outside the training vocabulary, but they generally do not determine semantic names for the detected unknowns. Furthermore, the unconstrained nature of the open world makes the desired detection granularity inherently ambiguous: depending on the task, an object may correspond to an entire entity or to one of its constituent parts.

\textbf{Open-Vocabulary Object Detection.}
Open-Vocabulary Object Detection (OVOD) takes a complementary approach by allowing models to detect categories specified through arbitrary textual queries, typically by leveraging vision-language representations
\cite{gu2021open,zhou2022detic,zhong2022regionclip,du2022detpro,feng2022promptdet,li2022glip}.
This enables recognition of categories beyond those explicitly annotated during detector training without requiring category-specific retraining
\cite{kuo2023fvlm,wu2023cora,wu2023baron,yao2023detclipv2,shi2023edadet}.
More recent approaches have further improved language-conditioned and open-vocabulary detection through stronger pretraining, self-training, and semantic modeling
\cite{cho2024decola,zhao2024sasdet,liu2024shine,yao2024detclipv3}.
However, in standard OVOD formulations, the vocabulary to be grounded is still externally specified
\cite{minderer2022owlvit,owlv2,gdino,yoloworld}.
Given a set of textual queries, an OVOD model can determine where the corresponding concepts occur, but it does not generally determine which concepts should be queried in the first place. Recent open-ended approaches such as GenerateU begin to relax this assumption by generating object names rather than relying exclusively on a predefined query set
\cite{lin2024generateu}.
Consequently, when the relevant objects are not known beforehand, standard OVOD systems still require a user or an upstream system to first construct the detection vocabulary.

\textbf{Multimodal Large Language Models.}
Recent multimodal Large Language Models (MLLMs) combine language reasoning with visual understanding and have demonstrated strong capabilities in tasks such as image understanding, visual question answering, and instruction-following over visual inputs
\cite{liu2023llava,bai2023qwenvl,wang2024qwen2vl,deitke2025molmo}.
Recent MLLMs have also incorporated explicit spatial grounding capabilities through coordinate generation, region representations, bounding boxes, masks, and point-based localization
\cite{peng2024kosmos2,chen2023shikra,you2024ferret,rasheed2024glamm,zhang2024llavagrounding,li2024groundinggpt,deitke2025molmo}.
Their ability to interpret an entire scene and produce semantically meaningful textual descriptions provides a natural mechanism for determining which visual concepts may be relevant to a particular image or task. However, general-purpose multimodal language models and specialized object detectors offer different strengths: MLLMs provide broad semantic reasoning and flexible language interaction, whereas dedicated detection models are designed for accurate and efficient spatial localization.

Our work combines these complementary capabilities through \textbf{M\textcolor{letterL}{L}LM-Guided \textcolor{letterA}{A}utonomous \textcolor{letterO}{O}bject \textcolor{letterD}{D}etection (\LAOD{})}. Rather than treating the detection vocabulary as fixed, as in conventional detectors, representing unseen objects only as ``unknown,'' as in OWOD, or requiring the vocabulary to be manually specified, as in standard OVOD, \LAOD{} uses an MLLM to autonomously construct an image-specific detection vocabulary. The resulting concepts are then grounded by an open-vocabulary detector. This decomposition separates the semantic decision of \emph{what} should be detected from the spatial task of determining \emph{where} those concepts occur. Moreover, because the vocabulary is generated dynamically from the image and task instruction, the detection scope can be adapted without retraining or manually enumerating the target categories.
\section{Methodology}

Our proposed \LAOD{} framework combines a multimodal large language model (MLLM) with an open-vocabulary object detector to perform autonomous object detection without requiring a predefined or externally specified category vocabulary at inference time. Given an input image and a task-level prompt, the MLLM first interprets the scene and autonomously determines a set of image-specific object categories to be detected. These generated categories are then provided to an open-vocabulary detector, which localizes the corresponding objects in the image.

This design separates two complementary capabilities: the MLLM determines \emph{what} concepts are relevant to the scene, while the open-vocabulary detector determines \emph{where} instances of those concepts occur. Importantly, the task instruction does not need to enumerate the target categories as in conventional open-vocabulary detection. Instead, it specifies the desired detection behavior at a higher level, while the detection vocabulary itself is constructed by the MLLM. As illustrated in Figure~\ref{fig:method}, \LAOD{} therefore consists of two sequential components: autonomous vocabulary generation and open-vocabulary object grounding.

\subsection{Autonomous Vocabulary Generation}
\label{sec:voc}

The first component of \LAOD{} is responsible for interpreting the visual input and constructing an image-specific detection vocabulary. We employ an MLLM that accepts both the input image and a textual prompt and produces a set of object class names that are relevant to the scene. Rather than receiving a fixed vocabulary or an explicitly enumerated list of target categories, the model determines which concepts should subsequently be queried by the detector.

For a given image $X$, an MLLM $L$ receives a prompt $p$ and generates an image-specific set of class names $C_g$:

\begin{equation}
C_g = L(X,p).
\end{equation}

By default, we use the prompt \textit{``Give me a list of objects that you see in this image. Just give the list of
objects comma seperated, don't explain anything. Do not list sky, street.''} In this setting, the MLLM is responsible for constructing the detection vocabulary directly from the visual content. The resulting vocabulary may vary across images, allowing \LAOD{} to adapt its detection targets without retraining or relying on a fixed category set.

The prompt $p$ also provides a natural mechanism for controlling the scope and granularity of detection. For example, a prompt requesting only ``main objects'' or ``primary entities'' may result in broad categories such as ``person'' and ``car,'' whereas a prompt requesting ``fine-grained parts'' may produce labels such as ``shirt,'' ``head,'' ``tire,'' or ``door.'' Similarly, task-specific prompts can restrict the generated vocabulary to concepts relevant to a particular application.

This flexibility addresses an inherent ambiguity in open-world detection: what constitutes an object of interest depends on the intended task and level of abstraction. In \LAOD{}, this choice does not need to be encoded through a fixed annotation vocabulary. Instead, the desired detection scope can be expressed through natural-language prompts, while the MLLM autonomously determines the corresponding image-specific categories.

\subsection{Open-Vocabulary Object Grounding}

Once the image-specific vocabulary $C_g$ has been generated, the second component localizes the corresponding concepts using an open-vocabulary object detector. Such detectors accept arbitrary textual category names as queries and identify image regions associated with those concepts.

The generated vocabulary is provided directly to the detector together with the original image. Formally, the detection stage is expressed as

\begin{equation}
D = O(X, C_g),
\label{eq:detection_output}
\end{equation}

where $O$ denotes the open-vocabulary detector and $D$ is the resulting set of detections. Each detection contains a bounding box, a predicted class name from $C_g$, and an associated detector confidence score. This score is produced by the open-vocabulary detector conditioned on the image and the supplied vocabulary and is used primarily to rank detections for evaluation. We do not interpret it as a globally calibrated probability over an unrestricted category space.

This two-stage formulation allows the semantic reasoning and spatial localization problems to be handled by models specialized for their respective roles. The MLLM performs scene interpretation and determines the detection vocabulary, while the open-vocabulary detector performs spatial grounding. Because the interface between the two components is simply a set of textual class names, \LAOD{} is modular and does not depend on a particular MLLM or open-vocabulary detector. We investigate this modularity experimentally by evaluating multiple combinations of both components. In our primary experiments, the labels generated by the MLLM are passed directly to the open-vocabulary detector without additional modification.

\section{Evaluation Metrics}

Evaluating \LAOD{} differs from conventional object detection because the
detection vocabulary is generated dynamically for each image rather than
selected from a fixed set of benchmark categories. A prediction may therefore
successfully localize an object while assigning it a semantically different
name from the ground-truth annotation, or conversely generate an appropriate
object name without accurately localizing the corresponding instance. We
therefore evaluate localization and semantic naming separately using
Class-Agnostic Average Precision (CAAP) and Semantic Naming Average Precision
(SNAP), respectively. However, these independent evaluations do not determine
whether an appropriate semantic name is assigned to the particular object that
was localized. To capture this joint requirement, we additionally introduce
Semantic Localization Average Precision (SemLoc-AP), which requires spatial
and semantic correctness with respect to the same ground-truth instance.
Together, CAAP, SNAP, and SemLoc-AP characterize localization quality,
semantic naming quality, and their joint correctness.

\subsection{Evaluation Protocol}

For an image $X$, let the detector output be
\begin{equation}
\mathcal{D}_X
=
\{(\hat{b}_i,\hat{l}_i,s_i)\}_{i=1}^{N_P^{(X)}},
\end{equation}
where $\hat{b}_i$ is a predicted bounding box, $\hat{l}_i$ is its predicted
class name, and $s_i$ is the confidence score produced by the open-vocabulary
detector conditioned on the supplied image-specific vocabulary. We use $s_i$
to rank predictions for evaluation and do not interpret it as a globally
calibrated probability over an unrestricted category space. Let the corresponding
ground-truth annotations be
\begin{equation}
\mathcal{G}_X
=
\{(b_j,l_j)\}_{j=1}^{N_{GT}^{(X)}},
\end{equation}
where $b_j$ and $l_j$ denote the ground-truth bounding box and class name,
respectively. Across an evaluation set $\mathcal{X}$, the total number of
ground-truth objects is
\begin{equation}
N_{GT}
=
\sum_{X\in\mathcal{X}} N_{GT}^{(X)}.
\end{equation}

Following the standard COCO evaluation convention, we retain at most the
top 100 predictions per image according to confidence. Predictions are
processed in descending order of $s_i$. Once a ground-truth object is assigned
to a prediction, it is marked as matched and cannot be assigned again,
resulting in one-to-one matching. Processing predictions in descending
confidence order ensures that, when multiple predictions compete for the same
ground-truth object, the more confident prediction is considered first. The
criterion used to determine whether a prediction and ground-truth object form
a valid match depends on the metric and is defined in the following
subsections. After image-level matching, predictions are pooled across the
evaluation set and ranked by confidence to construct precision--recall curves.
The same top-100 prediction set is used throughout the evaluation.

\subsection{Class-Agnostic Average Precision}

Class-Agnostic Average Precision (CAAP) measures the ability of a system to
detect and localize objects while disregarding their predicted class names.
It follows the standard COCO-style AP protocol after removing class identity
from the matching process. For a given Intersection over Union (IoU) threshold
$\theta$, each prediction $\hat{b}_i$ is compared with the currently unmatched
ground-truth boxes in the same image. Among the ground-truth boxes satisfying
the IoU criterion, the prediction is matched to the one with the highest IoU.
It is counted as a true positive if
\begin{equation}
\operatorname{IoU}(\hat{b}_i,b_j) \geq \theta.
\end{equation}
If no unmatched ground-truth box satisfies the threshold, the prediction is
counted as a false positive. This includes duplicate detections of an object
that has already been matched. Ground-truth objects that remain unmatched are
counted as false negatives. Predicted and ground-truth class names are ignored
entirely during this matching process.

At a detector confidence threshold $\gamma$, precision and recall are defined as
\begin{align}
P(\theta,\gamma)
&=
\frac{\operatorname{TP}(\theta,\gamma)}
{\operatorname{TP}(\theta,\gamma)+\operatorname{FP}(\theta,\gamma)},\\
R(\theta,\gamma)
&=
\frac{\operatorname{TP}(\theta,\gamma)}
{N_{GT}}.
\end{align}

For a fixed IoU threshold $\theta$, average precision is computed using the
standard COCO 101-point interpolated protocol. Let
\begin{equation}
\mathcal{R}
=
\{0,0.01,\ldots,1.00\}
\end{equation}
denote the 101 recall levels. We define the interpolated precision at recall
level $r$ as
\begin{equation}
\bar{P}_{\theta}(r)
=
\max_{\gamma:\,R(\theta,\gamma)\geq r}
P(\theta,\gamma),
\end{equation}
with $\bar{P}_{\theta}(r)=0$ when no operating point reaches recall $r$.
CAAP at IoU threshold $\theta$ is then
\begin{equation}
\operatorname{CAAP}(\theta)
=
\frac{1}{101}
\sum_{r\in\mathcal{R}}
\bar{P}_{\theta}(r).
\end{equation}

Our primary localization metric averages CAAP over the ten IoU thresholds
\begin{equation}
\Theta_{\mathrm{IoU}}
=
\{0.50,0.55,\ldots,0.95\},
\end{equation}
giving
\begin{equation}
\operatorname{CAAP}_{[.5:.95]}
=
\frac{1}{|\Theta_{\mathrm{IoU}}|}
\sum_{\theta\in\Theta_{\mathrm{IoU}}}
\operatorname{CAAP}(\theta).
\end{equation}

For additional analysis, we partition these thresholds into three
non-overlapping localization regimes:
\begin{align}
\Theta_{\mathrm{LO}} &= \{0.50,0.55,0.60\},\\
\Theta_{\mathrm{MI}} &= \{0.65,0.70,0.75,0.80\},\\
\Theta_{\mathrm{HI}} &= \{0.85,0.90,0.95\}.
\end{align}
The reported LO, MI, and HI scores are the mean CAAP values within the
corresponding subsets, whereas $\operatorname{CAAP}_{[.5:.95]}$ is the
unweighted mean over all ten IoU thresholds.

\subsection{Semantic Naming Average Precision}

CAAP intentionally ignores semantic labels and therefore does not measure
whether \LAOD{} assigns meaningful names to the objects it proposes. We
introduce Semantic Naming Average Precision (SNAP) to evaluate semantic naming
quality independently of spatial localization.

Each predicted class name $\hat{l}_i$ and ground-truth class name $l_j$ is
embedded using Sentence-BERT (\texttt{all-MiniLM-L6-v2}). We embed the
\emph{bare class names}, without adding a shared textual template. Let
\begin{equation}
E(\cdot): \mathrm{Text} \rightarrow \mathbb{R}^{d}
\end{equation}
denote the text encoder. Semantic similarity is measured using cosine
similarity:
\begin{equation}
\operatorname{sim}(\hat{l}_i,l_j)
=
\frac{
E(\hat{l}_i)^{\top}E(l_j)
}{
\|E(\hat{l}_i)\|_2\|E(l_j)\|_2
}.
\end{equation}

For a semantic similarity threshold $\tau$, predictions are processed in
descending detector-confidence order. Each predicted label $\hat{l}_i$ is
compared with the currently unmatched ground-truth labels from the same image.
Among these ground-truth objects, the prediction is matched to the one with
the highest semantic similarity if that similarity satisfies
\begin{equation}
\operatorname{sim}(\hat{l}_i,l_j) \geq \tau.
\end{equation}
If no unmatched ground-truth label satisfies the threshold, the prediction is
counted as a false positive. Once a ground-truth object is matched, it is
unavailable to subsequent predictions, and ground-truth objects that remain
unmatched are counted as false negatives. Bounding-box overlap is deliberately
excluded from SNAP so that semantic naming quality can be evaluated
independently of localization.

Importantly, the precision--recall curve for SNAP is ranked using the detector
confidence $s_i$, as in CAAP. The semantic similarity score is used only to
determine whether a prediction constitutes a correct semantic match at a fixed
threshold $\tau$.

At a detector confidence threshold $\gamma$, semantic precision and recall are
defined as
\begin{align}
P(\tau,\gamma)
&=
\frac{\operatorname{TP}(\tau,\gamma)}
{\operatorname{TP}(\tau,\gamma)+\operatorname{FP}(\tau,\gamma)},\\
R(\tau,\gamma)
&=
\frac{\operatorname{TP}(\tau,\gamma)}
{N_{GT}}.
\end{align}

For a fixed semantic similarity threshold $\tau$, we use the same COCO
101-point interpolated average-precision protocol. Let
\begin{equation}
\mathcal{R}
=
\{0,0.01,\ldots,1.00\}
\end{equation}
denote the 101 recall levels. We define the interpolated precision at recall
level $r$ as
\begin{equation}
\bar{P}_{\tau}(r)
=
\max_{\gamma:\,R(\tau,\gamma)\geq r}
P(\tau,\gamma),
\end{equation}
with $\bar{P}_{\tau}(r)=0$ when no operating point reaches recall $r$.
SNAP at semantic similarity threshold $\tau$ is then
\begin{equation}
\operatorname{SNAP}(\tau)
=
\frac{1}{101}
\sum_{r\in\mathcal{R}}
\bar{P}_{\tau}(r).
\end{equation}

We evaluate semantic matching over
\begin{equation}
\Theta_{\mathrm{sem}}
=
\{0.50,0.55,\ldots,0.95\},
\end{equation}
and report the primary SNAP score as
\begin{equation}
\operatorname{SNAP}_{[.5:.95]}
=
\frac{1}{|\Theta_{\mathrm{sem}}|}
\sum_{\tau\in\Theta_{\mathrm{sem}}}
\operatorname{SNAP}(\tau).
\end{equation}
We additionally report LO, MI, and HI semantic regimes using the same
non-overlapping threshold partition defined above.

\paragraph{Choice of semantic encoder.}
SNAP requires a text embedding space in which semantically unrelated category
names are sufficiently separated. We therefore use Sentence-BERT rather than
CLIP text embeddings. CLIP is primarily optimized for image--text alignment
rather than text--text semantic similarity and consequently assigns relatively
high cosine similarities even to distinct object categories. Using CLIP's
standard \textit{``a photo of \{\}''} text template, we find that across all
distinct pairs of the 80 COCO categories, the minimum cosine similarity is
$0.567$ and the median is $0.786$. Thus, at a semantic threshold of
$\tau=0.50$, every distinct COCO class pair satisfies the matching criterion,
preventing meaningful discrimination between related and unrelated names.

We further examine this behavior using a label-shuffling control in which
predicted label strings are randomly permuted across the dataset while the
bounding boxes and confidence scores remain unchanged. Under CLIP,
low-threshold SNAP is unchanged by this permutation, indicating that the
predicted labels themselves provide essentially no discriminative information
in this regime. In contrast, Sentence-BERT produces a clear separation between
the original and shuffled-label evaluations across the considered thresholds.
We therefore use Sentence-BERT with bare class names throughout our SNAP
evaluation.

\paragraph{Datasets without semantic category annotations.}
SNAP requires semantically meaningful ground-truth class names and is therefore
not applicable when all objects share a single generic annotation. In
particular, COCO-OOD assigns its out-of-distribution objects to a common
unknown-object category. We consequently do not report SNAP on COCO-OOD and
evaluate this benchmark using unknown-object localization metrics instead.

\subsection{Semantic Localization Average Precision}

CAAP and SNAP evaluate localization and semantic naming independently, but
neither metric determines whether an appropriate semantic name is assigned to
the particular object that was localized. For example, a prediction may
correctly localize a dog while being labeled ``cat,'' and SNAP may still match
that label to a different cat elsewhere in the image. We therefore introduce
Semantic Localization Average Precision (SemLoc-AP) to evaluate whether
spatial and semantic correctness hold for the same ground-truth instance.

SemLoc-AP combines the matching criteria used by CAAP and SNAP. For a
prediction $i$ and ground-truth object $j$, define the CAAP spatial match
indicator at IoU threshold $\theta$ as
\begin{equation}
m^{\mathrm{CAAP}}_{ij}(\theta)
=
\mathbb{I}
\left[
\operatorname{IoU}(\hat{b}_i,b_j)\geq\theta
\right],
\end{equation}
and the SNAP semantic match indicator at semantic threshold $\tau$ as
\begin{equation}
m^{\mathrm{SNAP}}_{ij}(\tau)
=
\mathbb{I}
\left[
\operatorname{sim}(\hat{l}_i,l_j)\geq\tau
\right].
\end{equation}
SemLoc-AP requires both conditions to hold for the same ground-truth object:
\begin{equation}
m^{\mathrm{SemLoc}}_{ij}(\theta,\tau)
=
m^{\mathrm{CAAP}}_{ij}(\theta)\,
m^{\mathrm{SNAP}}_{ij}(\tau).
\end{equation}

Predictions are processed using the same confidence ordering and one-to-one
matching protocol defined above. For each prediction, the eligible
ground-truth objects are the currently unmatched objects satisfying
$m^{\mathrm{SemLoc}}_{ij}(\theta,\tau)=1$. If multiple ground-truth objects
are eligible, the prediction is matched to the one with the highest IoU.
Thus, the semantic criterion determines eligibility, while spatial overlap
determines which eligible object the prediction localized.

For each threshold pair $(\theta,\tau)$, we denote by
$\operatorname{SemLocAP}(\theta,\tau)$ the average precision obtained under
this joint matching criterion using the same detector-confidence ranking and
COCO 101-point interpolated AP protocol defined for CAAP and SNAP.

Rather than coupling spatial and semantic thresholds along a diagonal, we
evaluate their full Cartesian product,
\begin{equation}
\theta \in \Theta_{\mathrm{IoU}},
\qquad
\tau \in \Theta_{\mathrm{sem}},
\end{equation}
because IoU and semantic-similarity thresholds have no principled one-to-one
correspondence. Our primary SemLoc-AP score is therefore

\begin{equation}
\operatorname{SemLoc\text{-}AP}
=
\frac{
\displaystyle
\sum_{\theta\in\Theta_{\mathrm{IoU}}}
\sum_{\tau\in\Theta_{\mathrm{sem}}}
\operatorname{SemLocAP}(\theta,\tau)
}{
|\Theta_{\mathrm{IoU}}|\,|\Theta_{\mathrm{sem}}|
}.
\end{equation}

With ten IoU thresholds and ten semantic-similarity thresholds, this averages
over 100 threshold pairs. SemLoc-AP is a pooled detection AP rather than a
per-class mean AP. CAAP, SNAP, and SemLoc-AP capture three distinct and
important aspects of performance: localization quality, semantic naming
quality, and joint semantic--spatial correctness. A low SemLoc-AP score alone
does not reveal whether the limiting factor is localization, semantic naming,
or both; we therefore report and analyze all three metrics together.

\begin{table}[t]
\centering
\caption{Illustrative interpretation of CAAP, SNAP, and SemLoc-AP. The three
metrics capture localization quality, semantic naming quality, and their
correct association, respectively.}
\label{tab:metric_interpretation}
\small
\begin{tabular}{p{0.39\linewidth}ccc}
\toprule
\textbf{Situation} & \textbf{CAAP} & \textbf{SNAP} & \textbf{SemLoc-AP} \\
\midrule
Good localization, poor naming
& High & Low & Low \\

Good naming, poor localization
& Low & High & Low \\

Both good and correctly associated
& High & High & High \\

Both good separately, but associated with different objects
& High & High & Low \\

Poor localization and naming
& Low & Low & Low \\
\bottomrule
\end{tabular}
\end{table}

Table~\ref{tab:metric_interpretation} summarizes how the three metrics should
be interpreted together. CAAP identifies whether the system successfully
localizes objects, while SNAP measures whether it produces semantically
appropriate object names. SemLoc-AP additionally measures whether the spatial
and semantic predictions are correct for the same ground-truth instance.
Consequently, CAAP and SNAP can both be high while SemLoc-AP remains low when
the localized objects and predicted names are not correctly associated. The
three metrics therefore provide distinct views of localization, naming, and
their joint correctness.
\section{Experiments}

\subsection{Experimental Setup}

We evaluate \LAOD{} across multiple multimodal large language models (MLLMs),
open-vocabulary object detectors, and benchmark datasets to study both
components of the proposed framework. Unless otherwise specified, the MLLM is
queried using the default instruction described in Section~\ref{sec:voc},
and the generated image-specific vocabulary is passed directly to the
open-vocabulary detector.

\paragraph{Models.}
We evaluate seven MLLMs for autonomous vocabulary generation:
Gemma3-4B~\cite{gemma3}, Gemma4-E2B, Gemma4-E4B, and
Gemma4-12B~\cite{gemma4}, Qwen2.5-VL-7B~\cite{qwen25vl},
Qwen3.5-9B~\cite{qwen35}, and InternVL3-8B~\cite{internvl3}.
For object grounding, we consider four open-vocabulary detectors:
YOLO-World~\cite{yoloworld}, Grounding DINO-T and Grounding
DINO-B~\cite{gdino}, and OWLv2~\cite{owlv2}. This produces
$7\times4=28$ MLLM--detector combinations in our main evaluation.
For each image, MLLM, and prompt combination, the generated vocabulary is
cached and reused across detectors, ensuring that all detectors are evaluated
using exactly the same set of generated class names.

\paragraph{Datasets.}
We evaluate on three datasets representing complementary detection settings.
COCO-Val~\cite{coco} provides a standard common-category benchmark for
analyzing the interaction between vocabulary generation and object grounding.
LVIS-minival~\cite{minival} contains a substantially larger and long-tailed
category vocabulary and is used to evaluate generalization to fine-grained and
less frequent object categories. COCO-OOD~\cite{unsniffer} contains objects
outside the standard COCO category set and is used to evaluate localization of
previously unknown objects.

For COCO-Val, we reserve 500 images for hyperparameter selection and evaluate
the primary results on the remaining 4,500 images. The LVIS-minival evaluation
contains 4,327 images, while the COCO-OOD evaluation contains 438 images.
The data splits are fixed before evaluation, and the held-out COCO images used
for hyperparameter selection are excluded from the reported COCO-Val test
results.

\paragraph{Detector configuration.}
Detector confidence thresholds are selected using the held-out data according
to the evaluation objective used for each benchmark: CAAP for COCO-Val and
LVIS, and U-AP for COCO-OOD. The resulting thresholds are $0.001$ for
YOLO-World, $0.03$ for both Grounding DINO variants, and $0.01$ for OWLv2.
For Grounding DINO, the text threshold is fixed at $0.3$. The selected
confidence threshold for each detector is consistent across the evaluated
MLLMs and datasets.

\begin{figure*}[t]
  \centering
   \includegraphics[width=0.9\linewidth]{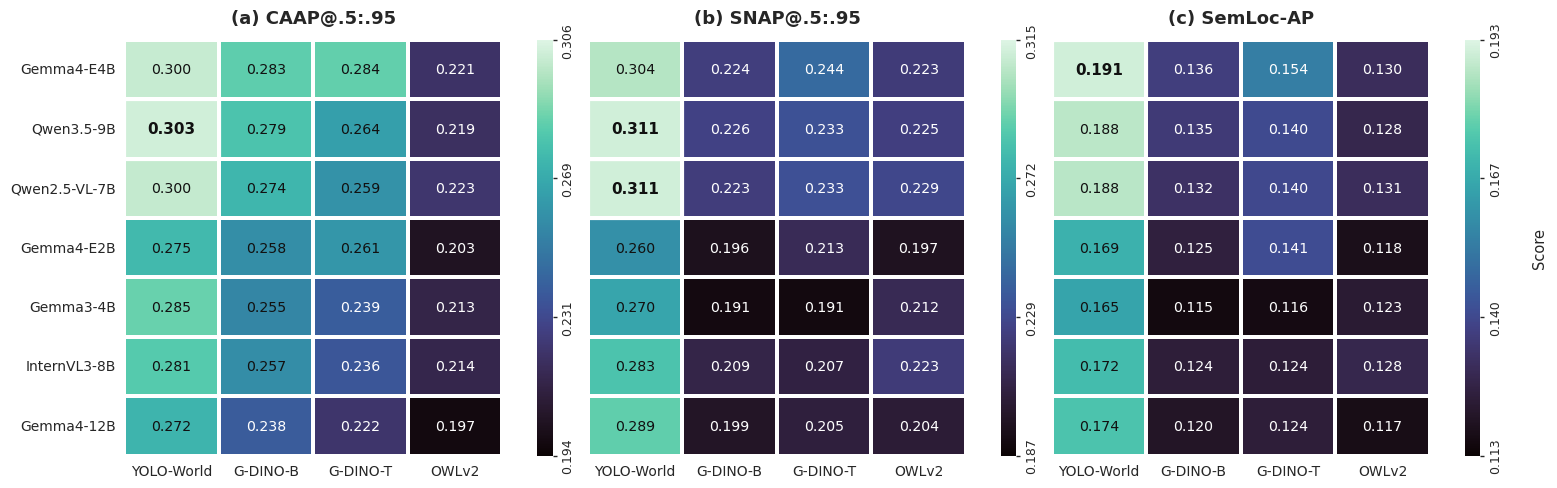}
   \caption{CAAP and SNAP on COCO-Val for all 28 combinations of seven MLLMs
   and four open-vocabulary detectors. Each cell reports the macro score
   averaged over thresholds from $0.50$ to $0.95$. Darker shading indicates
   higher performance.}
   \label{fig:grid_main}
\end{figure*}

\paragraph{Evaluation protocol.}
We report CAAP, SNAP, and SemLoc-AP using the protocol defined in the previous
section, with a maximum of 100 detections per image. CAAP and SNAP are macro
averaged over thresholds from $0.50$ to $0.95$, while SemLoc-AP is averaged
over the full Cartesian product of the corresponding ten IoU and ten semantic
similarity thresholds. For each open-vocabulary detector, we use a single
confidence threshold selected on held-out data and keep it fixed across
datasets and MLLMs. Detector confidence is used to rank predictions for AP
rather than interpreted as a globally calibrated probability. SNAP and
SemLoc-AP are reported only for datasets with meaningful semantic category
annotations. For COCO-OOD, where out-of-distribution objects share a generic
unknown-object label, we instead report unknown-object localization metrics
including U-AP, U-F1, U-PRE, and U-REC.

In addition to the primary MLLM--detector evaluation, we conduct three
additional analyses. First, we study how prompt formulation affects the
generated vocabulary and downstream detection performance through a prompt
ablation. Second, we compare \LAOD{} with MLLMs used directly as object
detectors to evaluate the benefit of separating semantic vocabulary generation
from spatial grounding. Finally, we provide qualitative examples illustrating
autonomous object discovery and controllable detection scope.


\subsection{Autonomous Detection on COCO}
\label{sec:coco_res}

We first evaluate the full \LAOD{} design on COCO-Val to examine how the
choice of MLLM and open-vocabulary detector jointly affects performance.
Figure~\ref{fig:grid_main} reports CAAP, SNAP, and SemLoc-AP for all
$7\times4=28$ MLLM--detector combinations. Because the generated vocabulary is
fixed for a given image and MLLM across detectors, differences within each row
directly reflect the effect of the grounding model.

The results show that \LAOD{} performs consistently across a diverse set of
MLLMs, while the choice of grounding model has a substantial effect on the
final detection quality. YOLO-World achieves the highest average CAAP
($0.2882$), followed by Grounding DINO-B ($0.2634$), Grounding DINO-T
($0.2522$), and OWLv2 ($0.2129$). The strongest individual configuration is
Qwen3.5-9B with YOLO-World, reaching a CAAP of $0.3030$. Qwen2.5-VL-7B and
Gemma4-E4B obtain nearly identical performance with the same detector, with
CAAP values of $0.3004$ and $0.3005$, respectively.

A similar trend appears for semantic naming, although the relative ordering of
the grounding models changes. YOLO-World again performs best, with an average
SNAP of $0.2899$, followed by Grounding DINO-T ($0.2179$), OWLv2 ($0.2159$),
and Grounding DINO-B ($0.2097$). The highest SNAP is $0.3110$, achieved by both
Qwen3.5-9B and Qwen2.5-VL-7B when paired with YOLO-World. These results
illustrate that localization and semantic naming are related but distinct
properties: a detector that produces stronger class-agnostic localization does
not necessarily preserve the same ranking under semantic evaluation.

SemLoc-AP further measures whether localization and semantic naming are correct
for the same ground-truth instance. YOLO-World again achieves the highest
average score ($0.1779$), followed by Grounding DINO-T ($0.1342$),
Grounding DINO-B ($0.1270$), and OWLv2 ($0.1249$). The strongest individual
configuration is Gemma4-E4B with YOLO-World, reaching a SemLoc-AP of $0.1911$.
The differences between the CAAP, SNAP, and SemLoc-AP rankings reinforce the
importance of evaluating localization, semantic naming, and their correct
association separately.

Across detectors, Gemma4-E4B obtains the highest average CAAP ($0.2722$) and
SemLoc-AP ($0.1531$), while Qwen2.5-VL-7B obtains the highest average SNAP
($0.2490$), with Gemma4-E4B ($0.2487$) and Qwen3.5-9B ($0.2486$) performing
nearly identically. We note that Gemma4-E4B is a distinct model variant rather
than a quantized version of Gemma4-12B. Although E4B performs better on COCO,
this ordering is not consistent across datasets: on LVIS
(Sec.~\ref{sec:lvis_res}), Gemma4-12B outperforms E4B, with a more pronounced
difference for SNAP and SemLoc-AP. This suggests that relative MLLM
performance is dataset-dependent rather than being determined by model size
alone.

More broadly, the variation across detector averages is larger than the
variation across MLLM averages for all three metrics. This indicates that
autonomous vocabulary generation is only one part of the overall detection
problem: the ability of the open-vocabulary detector to reliably ground the
generated concepts remains a major determinant of end-to-end performance. At
the same time, the strong performance obtained with several different MLLMs
and detectors demonstrates that \LAOD{} is not tied to a particular model
pair, supporting its modular formulation.

\subsection{Generalization to Long-Tail Categories}
\label{sec:lvis_res}

We next evaluate \LAOD{} on LVIS-minival to study whether the behavior observed
on COCO extends to a substantially larger and more fine-grained category space.
We evaluate Qwen3.5-9B, Gemma4-12B, and Gemma4-E4B with all four
open-vocabulary detectors on the 4,327-image LVIS-minival split, using the same
default prompting strategy and evaluation protocol as in the COCO experiments.

\begin{figure}[t]
  \centering
   \includegraphics[width=0.9\linewidth]{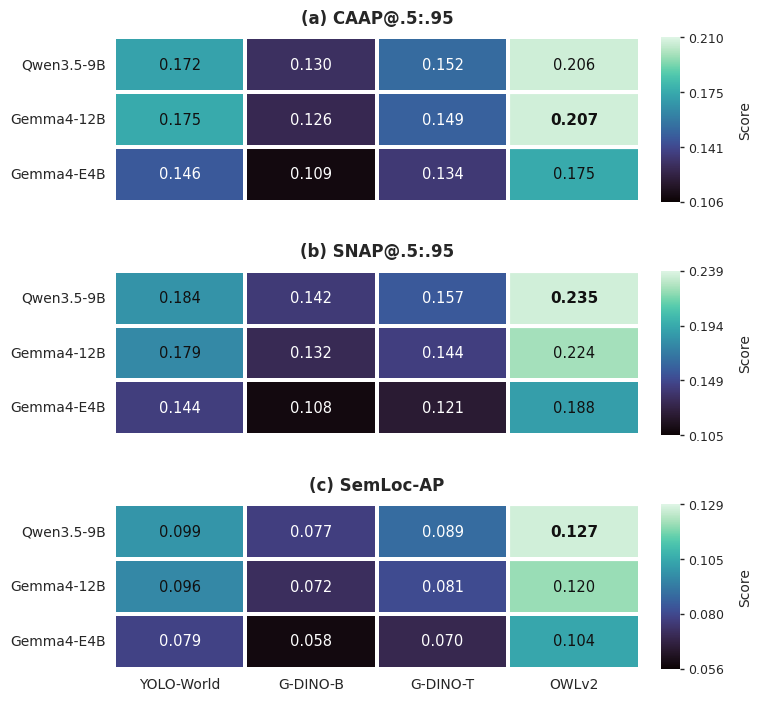}
   \caption{CAAP, SNAP, and SemLoc-AP on LVIS-minival for three MLLMs paired
   with four open-vocabulary detectors. The three grids are shown vertically
   for clarity. CAAP and SNAP report macro averages over thresholds from
   $0.50$ to $0.95$, while SemLoc-AP averages over the full Cartesian product
   of the corresponding spatial and semantic thresholds.}
   \label{fig:grid_lvis}
\end{figure}

As shown in Figure~\ref{fig:grid_lvis}, performance is lower than on COCO,
which is expected given the broader and more fine-grained long-tail category
space of LVIS. Nevertheless, \LAOD{} continues to produce meaningful
localization, semantic naming, and joint semantic--spatial performance across
all evaluated model combinations. The strongest CAAP is $0.2071$, obtained by
Gemma4-12B with OWLv2, while Qwen3.5-9B with OWLv2 achieves a nearly identical
$0.2064$. For SNAP, Qwen3.5-9B with OWLv2 performs best at $0.2352$, followed
by Gemma4-12B with OWLv2 at $0.2244$. The strongest SemLoc-AP is likewise
obtained by Qwen3.5-9B with OWLv2 at $0.1273$, followed by Gemma4-12B with
OWLv2 at $0.1204$.

A notable result is the change in the relative ranking of the grounding
models compared with COCO. On COCO, YOLO-World achieves the highest average
CAAP, SNAP, and SemLoc-AP across MLLMs. On LVIS, OWLv2 instead becomes the
strongest detector, reaching an average CAAP of $0.1963$, SNAP of $0.2159$,
and SemLoc-AP of $0.1172$, compared with $0.1645$, $0.1687$, and $0.0913$
for YOLO-World. Grounding DINO-T and Grounding DINO-B follow with average
CAAP values of $0.1451$ and $0.1215$, average SNAP values of $0.1408$ and
$0.1272$, and average SemLoc-AP values of $0.0797$ and $0.0689$,
respectively.

This shift in detector ranking supports the modular design of \LAOD{}.
The same MLLM-generated vocabulary can be paired with different grounding
models, and no single detector is uniformly strongest across datasets. While
YOLO-World is the strongest grounding model on the common-category COCO
benchmark, OWLv2 is substantially more effective on the long-tailed LVIS
setting across all three metrics. This suggests that the grounding component
can be selected according to the characteristics of the target domain without
changing the autonomous vocabulary-generation stage.

\subsection{Unknown-Object Detection on COCO-OOD}
\label{sec:ood_res}

We next evaluate \LAOD{} on COCO-OOD~\cite{unsniffer}, which is designed to
measure the ability to detect objects outside the standard COCO category set.
Unlike COCO and LVIS, COCO-OOD assigns out-of-distribution objects to a single
generic unknown category. Consequently, semantic naming and joint
semantic--spatial correctness cannot be meaningfully evaluated with SNAP or
SemLoc-AP on this benchmark. We therefore focus on class-agnostic localization
using CAAP and the standard unknown-object detection metrics U-AP, U-F1,
U-PRE, and U-REC.

\begin{figure}[t]
  \centering
  \includegraphics[width=0.9\linewidth]{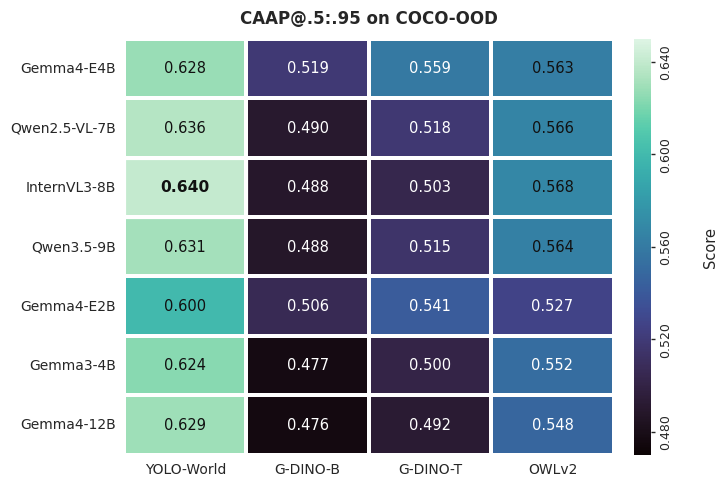}
  \caption{CAAP on COCO-OOD for all 28 combinations of seven MLLMs and four
  open-vocabulary detectors. Each cell reports the macro score averaged over
  IoU thresholds from $0.50$ to $0.95$.}
  \label{fig:grid_ood}
\end{figure}

Figure~\ref{fig:grid_ood} shows CAAP for all $7\times4=28$ MLLM--detector
combinations. YOLO-World consistently provides the strongest localization
performance across the evaluated MLLMs, with an average CAAP of $0.6267$,
compared with $0.5553$ for OWLv2, $0.5184$ for Grounding DINO-T, and $0.4920$
for Grounding DINO-B. The strongest individual configuration is InternVL3-8B
with YOLO-World, which reaches a CAAP of $0.6403$. Qwen2.5-VL-7B and
Qwen3.5-9B with YOLO-World achieve similarly strong scores of $0.6357$ and
$0.6306$, respectively. These results show that the image-specific vocabularies
generated by \LAOD{} can successfully guide open-vocabulary detectors toward
objects that fall outside the standard COCO category set.

In addition to CAAP, we compare \LAOD{} against existing open-vocabulary,
open-world, and unknown-object detection methods using the conventional
COCO-OOD metrics. Table~\ref{tab:uap-comparison} includes Grounding
DINO~\cite{gdino,undetr}, OW-DETR~\cite{owdetr}, VOS~\cite{vos},
UnSniffer~\cite{unsniffer}, and UN-DETR~\cite{undetr}. The standalone
Grounding DINO results are those reported by UN-DETR~\cite{undetr}, where
Grounding DINO is queried with the generic prompt ``unknown object.'' This
differs from Grounding DINO used within \LAOD{}, which receives the
image-specific vocabulary generated by the MLLM. Rather than selecting the
best \LAOD{} configuration specifically on COCO-OOD, we report our
\emph{main model}, Qwen3.5-9B with YOLO-World. This configuration is selected
from the COCO-Val experiments, where it obtains the highest CAAP and ties for
the highest SNAP, and is therefore fixed independently of the COCO-OOD
evaluation. Results for all 28 \LAOD{} MLLM--detector combinations are reported
in the supplementary material.

\begin{table}[t]
\centering
\caption{Unknown-object detection performance on COCO-OOD. For \LAOD{}, we
report the main Qwen3.5-9B + YOLO-World configuration selected from the
COCO-Val experiments rather than selecting a configuration specifically for
COCO-OOD. The standalone Grounding DINO result is reported by
UN-DETR~\cite{undetr} using the generic prompt ``unknown object.''}
\label{tab:uap-comparison}
\resizebox{\linewidth}{!}{
\begin{tabular}{lcccc}
\toprule
Method & U-AP & U-F1 & U-PRE & U-REC \\
\midrule
Grounding DINO~\cite{gdino,undetr}
    & 0.26 & 0.41 & \textbf{0.83} & 0.28 \\
OW-DETR~\cite{owdetr}
    & 0.03 & 0.05 & 0.03 & 0.38 \\
VOS~\cite{vos}
    & 0.20 & 0.31 & 0.29 & 0.34 \\
UnSniffer~\cite{unsniffer}
    & 0.45 & 0.47 & 0.43 & 0.53 \\
UN-DETR~\cite{undetr}
    & 0.47 & 0.54 & 0.54 & 0.55 \\
\midrule
\LAOD{} (Qwen3.5-9B + YOLO-World)
    & \textbf{0.769} & \textbf{0.720} & 0.787 & \textbf{0.663} \\
\bottomrule
\end{tabular}
}
\end{table}

As shown in Table~\ref{tab:uap-comparison}, the main \LAOD{} configuration
achieves a U-AP of $0.769$, substantially exceeding the strongest competing
method, UN-DETR~\cite{undetr}, at $0.47$. \LAOD{} also obtains the highest
U-F1 ($0.720$) and U-REC ($0.663$), while achieving the second-highest U-PRE
($0.787$). Grounding DINO reaches a slightly higher precision of $0.83$, but
with substantially lower recall at $0.28$, reflecting a more conservative
precision--recall trade-off. In contrast, \LAOD{} maintains high precision
while recovering a considerably larger fraction of unknown objects.

Taken together, the CAAP grid and unknown-object metrics demonstrate that the
ability of \LAOD{} to autonomously construct image-specific detection
vocabularies extends beyond common and long-tailed categories to objects that
are explicitly outside the standard COCO label space. Importantly, the strong
results in Table~\ref{tab:uap-comparison} are obtained with the same main model
selected on COCO-Val, rather than a configuration chosen to maximize
COCO-OOD performance.

\subsection{Open-Ended Detection Baselines}

The closest task-level competitor to \LAOD{} is
GenerateU~\cite{lin2024generateu}, which was developed and trained specifically
for open-ended object detection. GenerateU learns a unified detector through
region-language pretraining to localize image regions and assign them free-form
object names without requiring a predefined category vocabulary at inference
time. This makes it a direct comparison to our setting: both methods aim to
detect and semantically name objects when the target categories are not known
in advance. The key distinction is architectural. GenerateU jointly learns
region proposal, localization, and semantic generation within a task-specific
open-ended detector, whereas \LAOD{} decomposes the problem across pretrained
specialists: an MLLM determines \emph{what} concepts are relevant to the image,
and an open-vocabulary detector determines \emph{where} those concepts occur,
without task-specific training of the combined pipeline.

We evaluate the strongest released GenerateU checkpoint, using a Swin-L
backbone trained on Visual Genome and GRIT-5M. GenerateU produces multiple
beam hypotheses for each localized region, whereas our evaluation assumes one
semantic label per detection. We therefore use the first generated hypothesis
as the predicted object name. We preserve the generated free-form labels and
do not map them to the benchmark ontology. This distinction is important
because GenerateU's released evaluation score combines its localization score
with CLIP similarity to LVIS category embeddings and subsequently maps the
prediction to an LVIS category. To preserve the ontology-free setting, we
instead rank predictions using the preceding native
objectness--IoU-quality score, before this category-dependent remapping.
GenerateU emits 300 predictions per image; as with all other methods, we apply
$\mathrm{maxDets}=100$ once before evaluation.

We additionally consider a second natural alternative: using a
grounding-capable MLLM directly as the complete object detector. Modern MLLMs
can generate both object names and bounding-box coordinates, so this comparison
tests whether the semantic and spatial components of \LAOD{} should be
separated at all. We evaluate Qwen2.5-VL-7B~\cite{qwen25vl} and
InternVL3-8B~\cite{internvl3} both as direct detectors and as the
vocabulary-generation component of \LAOD{} with
YOLO-World~\cite{yoloworld}. For direct detection, the models are prompted to
return object labels and bounding boxes in JSON format and use greedy decoding
with at most 1024 generated tokens. Since they do not provide native
detection-confidence scores, predictions are ranked using the mean token
log-probability associated with the generated bounding-box coordinates.

\begin{table}[t]
\centering
\caption{Comparison with GenerateU, a task-specific open-ended detector, and
direct MLLM detection. For the ``\LAOD{} detector avg.'' rows, each range spans
the four OVOD-specific detector averages, where each detector is averaged over
the evaluated MLLMs. SNAP and SemLoc-AP are not reported on COCO-OOD because
all out-of-distribution objects share a single generic unknown label.}
\label{tab:direct_vlm}
\resizebox{\linewidth}{!}{
\begin{tabular}{llcccc}
\toprule
Dataset & Method & CAAP & SNAP & SemLoc-AP & U-AP \\
\midrule

\multirow{6}{*}{COCO-Val}
& GenerateU (Swin-L, VG+GRIT5M)
    & 0.1976 & 0.1470 & 0.0821 & -- \\
& \LAOD{} detector avg. range
    & 0.2129--0.2882 & 0.2097--0.2899 & 0.1249--0.1779 & -- \\
& Qwen2.5-VL-7B Direct
    & 0.1422 & 0.0842 & 0.0359 & -- \\
& \LAOD{} (Qwen2.5-VL-7B + YOLO-World)
    & \textbf{0.3004} & \textbf{0.3110} & \textbf{0.1875} & -- \\
& InternVL3-8B Direct
    & 0.1329 & 0.2072 & 0.0841 & -- \\
& \LAOD{} (InternVL3-8B + YOLO-World)
    & \textbf{0.2815} & \textbf{0.2835} & \textbf{0.1716} & -- \\

\midrule

\multirow{2}{*}{LVIS-minival}
& GenerateU (Swin-L, VG+GRIT5M)
    & 0.0980 & 0.0984 & 0.0496 & -- \\
& \LAOD{} detector avg. range
    & 0.1215--0.1963 & 0.1272--0.2159 & 0.0689--0.1172 & -- \\

\midrule

\multirow{4}{*}{COCO-OOD}
& Qwen2.5-VL-7B Direct
    & 0.4108 & -- & -- & 0.553 \\
& \LAOD{} (Qwen2.5-VL-7B + YOLO-World)
    & \textbf{0.6357} & -- & -- & \textbf{0.772} \\
& InternVL3-8B Direct
    & 0.3748 & -- & -- & 0.496 \\
& \LAOD{} (InternVL3-8B + YOLO-World)
    & \textbf{0.6403} & -- & -- & \textbf{0.776} \\

\bottomrule
\end{tabular}
}
\end{table}

Table~\ref{tab:direct_vlm} shows that \LAOD{} consistently outperforms
GenerateU under the same ontology-free evaluation protocol. On COCO-Val,
GenerateU obtains $0.1976$ CAAP, $0.1470$ SNAP, and $0.0821$ SemLoc-AP.
In comparison, the four \LAOD{} detector averages range from
$0.2129$--$0.2882$ CAAP, $0.2097$--$0.2899$ SNAP, and
$0.1249$--$0.1779$ SemLoc-AP. Thus, GenerateU remains below even the lowest
\LAOD{} detector average on all three metrics.

The same result holds on the substantially larger and more fine-grained LVIS
ontology. GenerateU reaches $0.0980$ CAAP, $0.0984$ SNAP, and $0.0496$
SemLoc-AP, whereas the corresponding \LAOD{} detector averages span
$0.1215$--$0.1963$, $0.1272$--$0.2159$, and $0.0689$--$0.1172$,
respectively. This comparison is particularly informative because GenerateU is
explicitly trained for open-ended detection, while \LAOD{} does not train a new
end-to-end detector for this task. Instead, it composes pretrained models
through a textual vocabulary interface. The consistent advantage across both
datasets suggests that decomposing semantic concept discovery and spatial
grounding across specialized components is effective even relative to a
dedicated open-ended detection model.

The GenerateU outputs also illustrate some of the challenges of solving both
parts of the problem within a unified detector. The released model produces
exactly 300 candidate detections per image without score-threshold filtering or
NMS in its default inference path. Among the top 100 predictions retained by
our common evaluation protocol, approximately $55\%$ of generated names are
duplicates within an image. Its vocabulary is also highly open-ended:
frequently generated concepts include scene elements and object parts such as
\emph{wall}, \emph{window}, \emph{ground}, \emph{head}, and \emph{leg}; on
COCO, approximately $10.7\%$ of predictions contain common part-level terms.
These predictions are not necessarily semantically invalid, but many do not
correspond to the object-level annotations used by COCO and LVIS. In contrast,
\LAOD{} delegates dense instance enumeration and localization to an
open-vocabulary detector while using the MLLM primarily for scene-level
semantic discovery.

The direct-MLLM experiments provide complementary evidence for the same
decomposition. On COCO-Val, Qwen2.5-VL-7B improves from $0.1422$ to $0.3004$
CAAP when moved from direct detection into \LAOD{}, while InternVL3-8B improves
from $0.1329$ to $0.2815$. Corresponding gains are also observed for SNAP and
SemLoc-AP. On COCO-OOD, U-AP increases from $0.553$ to $0.772$ for
Qwen2.5-VL-7B and from $0.496$ to $0.776$ for InternVL3-8B.

These direct outputs are also considerably sparser: Qwen2.5-VL-7B and
InternVL3-8B produce approximately $2.5$ and $4.5$ detections per image on
COCO-Val, compared with approximately $57.9$ and $59.1$ detections per image
when the same MLLMs generate the vocabulary for YOLO-World within \LAOD{}.
The direct MLLMs can therefore localize objects, but their outputs behave more
like sparse semantic descriptions of salient objects than dense
instance-level detections.

Taken together, the GenerateU and direct-MLLM comparisons support the same
conclusion from two complementary directions. A dedicated open-ended detector
can jointly generate regions and semantic names, and modern MLLMs can directly
produce object coordinates. However, in our experiments, the strongest results
are obtained by separating semantic concept discovery from spatial grounding:
the MLLM determines \emph{what} concepts are relevant to the scene, while the
specialized open-vocabulary detector performs dense enumeration, localization,
and confidence-ranked prediction of \emph{where} those concepts occur.

\subsection{Prompt Sensitivity and Vocabulary Control}

We finally examine how sensitive \LAOD{} is to the prompt used for
autonomous vocabulary generation. We compare three prompts. The
\emph{Default} prompt is the generic prompt used throughout all main
experiments. The \emph{Minimal} prompt removes most formatting and exclusion
constraints, while the \emph{COCO-specific} prompt introduces additional
constraints that encourage coarse, singular object names and suppress
infrastructural objects. The three prompt variants are summarized in
Table~\ref{tab:prompt-definitions}.

\begin{table}[t]
\centering
\caption{Prompt variants used in the sensitivity analysis. All variants use the
same system prompt; only the user prompt changes.}
\label{tab:prompt-definitions}
\resizebox{\linewidth}{!}{
\begin{tabular}{lp{0.72\linewidth}}
\toprule
Prompt & User prompt \\
\midrule
Default &
Give me a list of objects that you see in this image. Just give the list of
objects comma seperated, don't explain anything. Do not list sky, street. \\

Minimal &
List the objects that you see in this image. \\

COCO-specific &
Give me a list of objects that you see in this image. Just give the list of
objects comma seperated, don't explain anything. Write only the general name
of the objects. Do not list infrastructural objects. Don't use plural, use
singular object names. \\
\bottomrule
\end{tabular}
}
\end{table}

To make the cross-dataset comparison directly comparable, we summarize results
using the three MLLMs evaluated under all three prompts on both COCO-Val and
LVIS-minival: Gemma4-12B, Gemma4-E4B, and Qwen3.5-9B. Restricting the summary
to this common set ensures that differences between datasets reflect prompt
behavior rather than differences in the evaluated MLLM pool. Each result in
Table~\ref{tab:prompt-ablation} is therefore averaged over the same three
MLLMs, with YOLO-World fixed as the grounding model.

\begin{table}[t]
\centering
\caption{Prompt sensitivity on 500-image subsets of COCO-Val and
LVIS-minival. Scores are averaged over the three MLLMs common to both
experiments: Gemma4-12B, Gemma4-E4B, and Qwen3.5-9B. The final column reports
the mean per-MLLM relative CAAP change with respect to the Default prompt.}
\label{tab:prompt-ablation}
\resizebox{\linewidth}{!}{
\begin{tabular}{llcccc}
\toprule
Dataset & Prompt & CAAP & SNAP & SemLoc-AP & $\Delta$CAAP vs.\ Default \\
\midrule

\multirow{3}{*}{COCO-Val}
& Default       & 0.2857 & 0.2955 & 0.1740 & -- \\
& Minimal       & 0.2855 & 0.2853 & 0.1707 & $+0.0\%$ \\
& COCO-specific & \textbf{0.3257} & \textbf{0.4150} & \textbf{0.2204} & \textbf{$+14.3\%$} \\

\midrule

\multirow{3}{*}{LVIS-minival}
& Default       & \textbf{0.1737} & \textbf{0.1821} & \textbf{0.1058} & -- \\
& Minimal       & 0.1718 & 0.1745 & 0.1026 & $-1.2\%$ \\
& COCO-specific & 0.1514 & 0.1651 & 0.0983 & \textbf{$-13.1\%$} \\

\bottomrule
\end{tabular}
}
\end{table}

Two observations emerge from Table~\ref{tab:prompt-ablation}. First, ordinary
changes in prompt wording have little effect. Replacing the Default prompt
with the much shorter Minimal prompt leaves COCO performance essentially
unchanged and reduces LVIS CAAP by only $1.2\%$. SNAP and SemLoc-AP show the
same general stability under this change. This suggests that \LAOD{} does not
rely strongly on the exact phrasing of a generic object-listing prompt.

Second, benchmark-specific prompt specialization has a much larger effect, but
the effect does not transfer across datasets. The COCO-specific prompt improves
CAAP by an average of $14.3\%$ over the Default prompt on COCO-Val and
outperforms it for all three common MLLMs. However, the same prompt reduces
CAAP by $13.1\%$ on LVIS-minival and underperforms the Default prompt for all
three models. The same pattern is visible in SNAP and SemLoc-AP. On COCO,
the specialized prompt increases the mean SNAP from $0.2955$ to $0.4150$ and
SemLoc-AP from $0.1740$ to $0.2204$. On LVIS, it reduces SNAP from $0.1821$
to $0.1651$ and SemLoc-AP from $0.1058$ to $0.0983$.

This behavior follows directly from the constraints introduced by the
COCO-specific prompt. Requests for general object names, suppression of
infrastructural objects, and singular labels align well with the relatively
coarse COCO ontology. In contrast, LVIS contains substantially finer-grained
categories and explicitly annotates concepts that such constraints may
suppress. The specialized prompt therefore improves agreement with the COCO
annotation convention rather than providing a general improvement in visual
recognition.

For systematic benchmarking demonstrated in Sections~\ref{sec:coco_res},
\ref{sec:lvis_res}, and~\ref{sec:ood_res}, we use the same Default prompt
across all datasets so that performance differences reflect the underlying
models and datasets rather than benchmark-specific prompt tuning. This
controlled setting provides a consistent basis for evaluating autonomous
vocabulary generation and comparing MLLM--detector configurations. At the
same time, the prompt ablation and the model combinations evaluated in the
main experiments highlight an important property of \LAOD{}: the framework is
highly configurable. Its vocabulary-generation prompt, MLLM, and grounding
detector can be adapted to the characteristics and requirements of a
particular scene or application. The benchmark results should therefore be
viewed as a systematic evaluation under a common configuration, rather than as
the performance limit of a scene-specific \LAOD{} system.

\subsection{Qualitative Results}

\newcommand{\qualimg}[1]{%
    \includegraphics[
        width=\linewidth,
        height=3.0cm,
        keepaspectratio
    ]{#1}%
}

\begin{figure*}[t]
    \centering

    \begin{subfigure}[t]{0.32\linewidth}
        \centering
        \qualimg{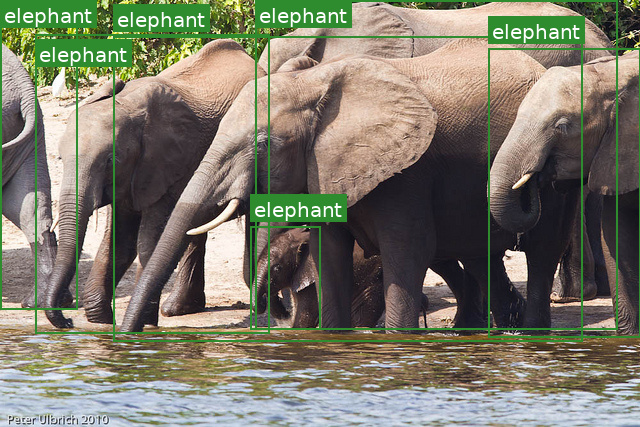}\\[-1mm]
        {\scriptsize Ground truth}
    \end{subfigure}
    \hfill
    \begin{subfigure}[t]{0.32\linewidth}
        \centering
        \qualimg{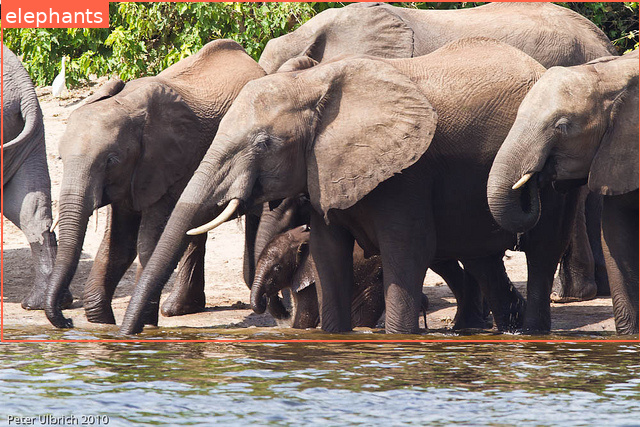}\\[-1mm]
        {\scriptsize Direct Qwen2.5-VL-7B}
    \end{subfigure}
    \hfill
    \begin{subfigure}[t]{0.32\linewidth}
        \centering
        \qualimg{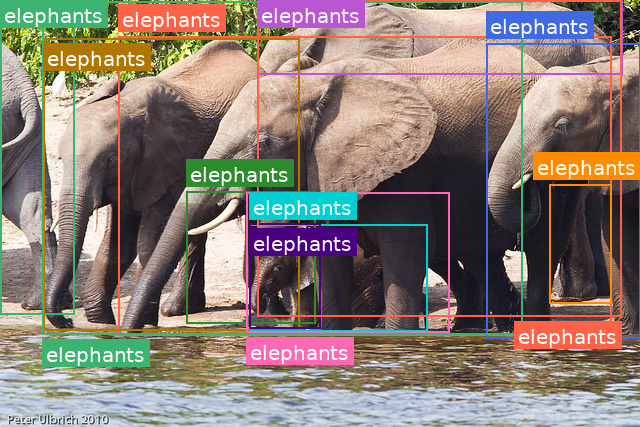}\\[-1mm]
        {\scriptsize \LAOD{} with YOLO-World}
    \end{subfigure}

    \vspace{1mm}

    {\scriptsize\textbf{(a)} Sparse instance enumeration}

    \par\vspace{2mm}

    \begin{subfigure}[t]{0.32\linewidth}
        \centering
        \qualimg{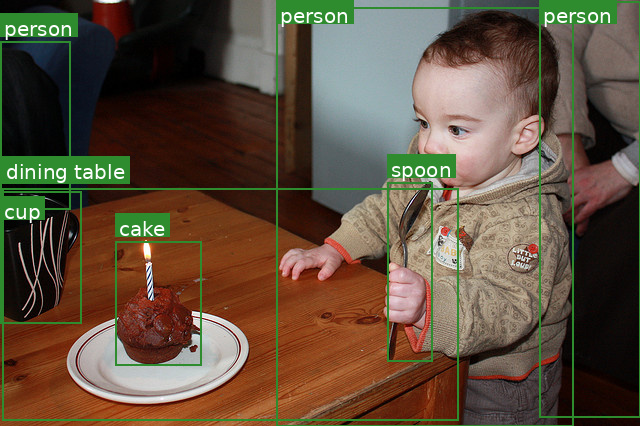}\\[-1mm]
        {\scriptsize Ground truth}
    \end{subfigure}
    \hfill
    \begin{subfigure}[t]{0.32\linewidth}
        \centering
        \qualimg{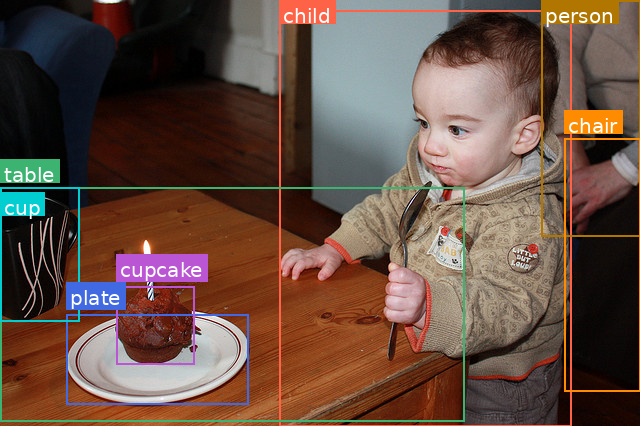}\\[-1mm]
        {\scriptsize Direct InternVL3-8B}
    \end{subfigure}
    \hfill
    \begin{subfigure}[t]{0.32\linewidth}
        \centering
        \qualimg{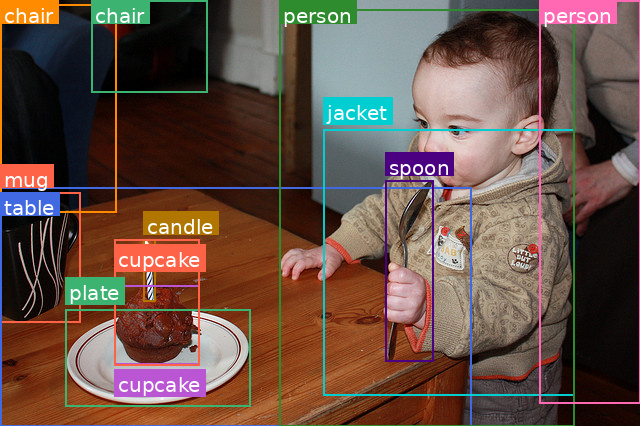}\\[-1mm]
        {\scriptsize \LAOD{} with YOLO-World}
    \end{subfigure}

    \vspace{1mm}

    {\scriptsize\textbf{(b)} Missed and spurious object detections}

    \par\vspace{2mm}

    \begin{subfigure}[t]{0.32\linewidth}
        \centering
        \qualimg{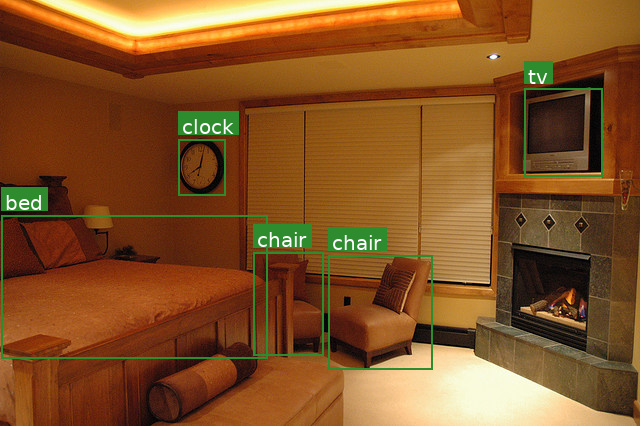}\\[-1mm]
        {\scriptsize Ground truth}
    \end{subfigure}
    \hfill
    \begin{subfigure}[t]{0.32\linewidth}
        \centering
        \qualimg{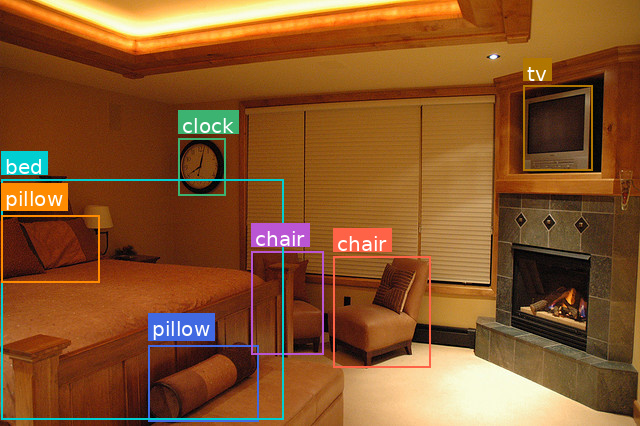}\\[-1mm]
        {\scriptsize Direct Qwen2.5-VL-7B}
    \end{subfigure}
    \hfill
    \begin{subfigure}[t]{0.32\linewidth}
        \centering
        \qualimg{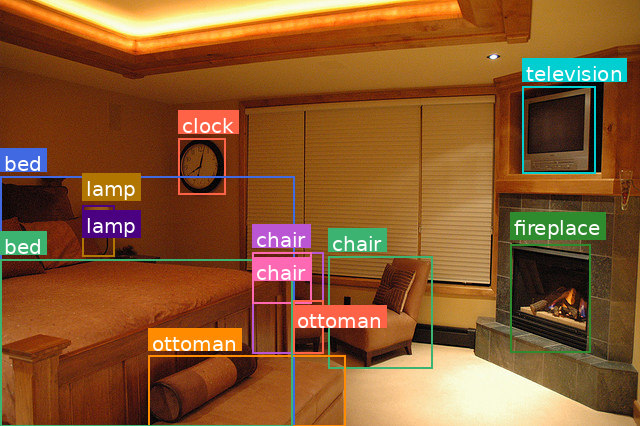}\\[-1mm]
        {\scriptsize \LAOD{} with YOLO-World}
    \end{subfigure}

    \vspace{1mm}

    {\scriptsize\textbf{(c)} Successful direct detection}

    \caption{Qualitative comparison between direct MLLM detection and the
    two-stage \LAOD{} formulation on COCO-Val. In each row, the same MLLM is
    used either to directly generate object labels and bounding boxes (middle)
    or to generate the image-specific vocabulary subsequently grounded by
    YOLO-World (right). In (a), direct Qwen2.5-VL-7B produces a single
    prediction for a group of elephants rather than enumerating the individual
    instances, while the corresponding \LAOD{} configuration localizes them
    separately. In (b), InternVL3-8B produces a generally plausible set of object detections,
    but misses an annotated spoon and introduces a spurious chair detection,
    illustrating that direct MLLM detection can still omit relevant instances and
    introduce unsupported objects even when the overall scene interpretation is
    reasonable. Panel (c) shows a case
    in which direct MLLM detection succeeds, demonstrating that the aggregate
    performance gap does not imply that MLLMs are incapable of generating
    accurate object detections. For readability, the \LAOD{} panels show only
    the highest-ranked predictions; quantitative evaluation uses the complete
    prediction set subject to the common maximum of 100 detections per image.}
    \label{fig:direct_mllm_qualitative}
\end{figure*}

Figure~\ref{fig:direct_mllm_qualitative} provides representative qualitative
examples comparing direct MLLM detection with the two-stage \LAOD{} design.
The examples show that \LAOD{} can recover annotated objects with denser and
more accurate spatial grounding, while direct MLLM detection may produce sparse
instance enumeration or inaccurate bounding boxes. At the same time, the
successful direct-detection example illustrates that MLLMs are capable of
generating valid object detections; the advantage of \LAOD{} arises from
delegating dense spatial grounding to a specialized detector rather than from
an inability of MLLMs to localize objects.

The examples also illustrate a broader property of autonomous vocabulary
generation: predicted object names need not exactly match a fixed benchmark
ontology to be semantically meaningful. Additional qualitative examples across
COCO-Val, LVIS-minival, and COCO-OOD are provided in
Appendix~\ref{app:additional_qualitative}.
\section{Limitations and Future Directions}

Despite its ability to construct image-specific detection vocabularies without
requiring users to enumerate object categories, \LAOD{} has several limitations.
First, vocabulary generation introduces additional computational cost because an
MLLM must process each image before object grounding. This overhead may limit
deployment in resource-constrained or high-throughput applications. Prior work
has shown that smaller, distilled, or task-specific language models can provide
substantial efficiency gains in specialized settings
\cite{small1,small2,small3,mumcu2024fast}. Future work could investigate whether
more compact multimodal models can preserve the vocabulary-generation capability
of \LAOD{} while reducing latency and memory requirements.

Second, the quality of the final detections depends on both components of the
framework. The MLLM determines \emph{what} concepts should be detected, whereas
the open-vocabulary detector determines \emph{where} those concepts occur. Our
experiments show that the strongest grounding model is dataset-dependent:
YOLO-World performs best on COCO and COCO-OOD, while OWLv2 performs
substantially better on the long-tailed LVIS setting. Although the modular
interface allows the grounding component to be exchanged without modifying the
vocabulary-generation stage, no single detector is uniformly optimal. Future
work could therefore explore automatic detector selection, detector ensembles,
or tighter interaction between vocabulary generation and grounding.

Third, autonomous vocabulary generation is susceptible to hallucinations by the
MLLM. The generated vocabulary may contain object concepts that are not
visually supported by the image, while relevant objects may also be omitted.
Because the generated class names are passed directly to the open-vocabulary
detector, errors introduced during vocabulary generation can propagate to the
grounding stage. A hallucinated concept may lead to spurious grounding
predictions, whereas an omitted concept cannot be recovered by the detector
because it was never included in the image-specific vocabulary. Future work
could investigate uncertainty-aware vocabulary generation, semantic
self-verification, or feedback between the MLLM and grounding model to reduce
such errors without requiring a predefined category set.

A further limitation concerns the relationship between generated concepts and
dataset annotation conventions. Our prompt-sensitivity experiments show that
ordinary changes in generic prompt wording have relatively little effect, but
benchmark-specific prompts can substantially alter performance. In particular,
a prompt specialized toward the COCO ontology improves performance on COCO
while reducing performance on LVIS. This indicates that prompt engineering can
increase agreement with a particular benchmark without necessarily improving
general object understanding. Developing methods that produce appropriately
granular object concepts without relying on benchmark-specific prompt design
remains an important direction.

Finally, existing object-detection benchmarks only partially capture the
capabilities of autonomous vocabulary generation. Conventional datasets are
defined by fixed annotation ontologies, while COCO-OOD evaluates unknown-object
localization using a single generic unknown label. Such settings do not fully
measure the quality or usefulness of dynamically generated semantic concepts.
Future benchmarks could evaluate localization together with semantic naming,
correct semantic--spatial association, concept granularity, and generalization
across annotation schemes.

\section{Conclusion}

We introduced \LAOD{}, an
M\textcolor{letterL}{L}LM-Guided
\textcolor{letterA}{A}utonomous
\textcolor{letterO}{O}bject
\textcolor{letterD}{D}etection framework
that separates semantic concept generation from spatial grounding. Given an
image and a generic prompt, an MLLM constructs an image-specific detection
vocabulary, which is then localized by an open-vocabulary detector. This removes
the need for users to enumerate target categories at inference time while
preserving a simple modular interface between the two components.

Across COCO-Val, LVIS-minival, and COCO-OOD, \LAOD{} demonstrates strong
localization, semantic naming, and joint semantic--spatial performance under
CAAP, SNAP, and SemLoc-AP, while also achieving strong unknown-object detection
results on COCO-OOD. Our experiments further show that separating semantic
discovery from spatial grounding is substantially more effective than using the
same MLLMs directly as object detectors, that the strongest grounding model
depends on the target dataset, and that benchmark-specific prompt engineering
can yield large but non-transferable gains. Overall, the results support a
modular view of autonomous object detection in which MLLMs determine
\emph{what} should be detected and open-vocabulary detectors determine
\emph{where} those concepts occur.

\bibliographystyle{plainnat}
\bibliography{references}

\clearpage
\onecolumn
\startcontents[appendix]

\begingroup
\normalsize
\setlength{\parskip}{0pt}
\setlength{\baselineskip}{11.5pt}
\vspace{-2mm}

\section*{Appendix Contents}
\printcontents[appendix]{}{1}{
    \setcounter{tocdepth}{2}
}

\endgroup

\clearpage
\twocolumn
\appendix

\section{Threshold-Wise CAAP, SNAP, and SemLoc-AP Results}
\label{app:interval_results}

The main paper reports CAAP and SNAP as macro scores averaged over thresholds
from $0.50$ to $0.95$, while SemLoc-AP averages over the full Cartesian
product of the corresponding spatial and semantic thresholds. Here, we provide
a more detailed view of the behavior of all three metrics.

For both CAAP and SNAP, we use the same disjoint low (LO), medium (MI), and
high (HI) threshold ranges:
\begin{equation}
\begin{aligned}
\mathcal{T}_{\mathrm{LO}}
    &= \{0.50,0.55,0.60\}, \\
\mathcal{T}_{\mathrm{MI}}
    &= \{0.65,0.70,0.75,0.80\}, \\
\mathcal{T}_{\mathrm{HI}}
    &= \{0.85,0.90,0.95\}.
\end{aligned}
\end{equation}
For CAAP, these correspond to increasingly strict IoU thresholds; for SNAP,
they correspond to increasingly strict semantic-similarity thresholds. Each
interval score is the mean AP over the thresholds contained in that interval.

Importantly, the macro score reported throughout the main paper is the
unweighted mean over all ten individual thresholds, rather than the mean of
the three interval scores. If $M_{\mathrm{LO}}$, $M_{\mathrm{MI}}$, and
$M_{\mathrm{HI}}$ denote the corresponding interval averages for either
CAAP or SNAP, then
\begin{equation}
M_{\mathrm{macro}}
=
\frac{
3M_{\mathrm{LO}}
+
4M_{\mathrm{MI}}
+
3M_{\mathrm{HI}}
}{10}.
\end{equation}
The weighting reflects the fact that the MI interval contains four thresholds,
whereas LO and HI each contain three.

SemLoc-AP depends jointly on an IoU threshold $\theta$ and a semantic
similarity threshold $\tau$ and therefore does not have a one-dimensional
LO/MI/HI decomposition. Instead, we summarize its threshold behavior using
two marginals. For a fixed IoU threshold, we average over all semantic
thresholds,
\begin{equation}
S_{\mathrm{IoU}}(\theta)
=
\frac{1}{|\Theta_{\mathrm{sem}}|}
\sum_{\tau\in\Theta_{\mathrm{sem}}}
\operatorname{SemLocAP}(\theta,\tau),
\end{equation}
and for a fixed semantic threshold, we average over all IoU thresholds,
\begin{equation}
S_{\mathrm{sem}}(\tau)
=
\frac{1}{|\Theta_{\mathrm{IoU}}|}
\sum_{\theta\in\Theta_{\mathrm{IoU}}}
\operatorname{SemLocAP}(\theta,\tau).
\end{equation}
The overall SemLoc-AP score is equivalently the mean of either set of ten
marginal values. All results below use the same prediction sets, confidence
ordering, one-to-one matching procedure, and maximum of 100 detections per
image as the main experiments.

\subsection{COCO-Val}

Tables~\ref{tab:app_coco_caap_intervals} and
\ref{tab:app_coco_snap_intervals} report the complete CAAP and SNAP
breakdowns on COCO-Val. Performance generally decreases as the localization
or semantic matching criterion becomes more stringent, while the relative
behavior observed in the macro results is largely maintained across threshold
ranges. Table~\ref{tab:app_coco_semloc_marginals} additionally reports the
SemLoc-AP marginals averaged across all 28 MLLM--detector combinations.

\begin{table*}[t]
\centering
\caption{Threshold-wise CAAP on COCO-Val. LO, MI, and HI denote the mean
AP over IoU thresholds $\{0.50,0.55,0.60\}$,
$\{0.65,0.70,0.75,0.80\}$, and $\{0.85,0.90,0.95\}$, respectively.
Macro is averaged over all ten IoU thresholds.}
\label{tab:app_coco_caap_intervals}
\scriptsize
\setlength{\tabcolsep}{5pt}
\renewcommand{\arraystretch}{0.93}
\begin{tabular}{llcccc}
\toprule
MLLM & Detector & LO & MI & HI & Macro \\
\midrule
Gemma4-E4B
    & YOLO-World & 0.4027 & 0.3354 & 0.1517 & 0.3005 \\
    & G-DINO-B   & 0.3762 & 0.3197 & 0.1419 & 0.2833 \\
    & G-DINO-T   & 0.3733 & 0.3183 & 0.1496 & 0.2842 \\
    & OWLv2      & 0.3094 & 0.2496 & 0.0945 & 0.2210 \\
\midrule
Qwen3.5-9B
    & YOLO-World & 0.4098 & 0.3389 & 0.1484 & 0.3030 \\
    & G-DINO-B   & 0.3762 & 0.3157 & 0.1316 & 0.2786 \\
    & G-DINO-T   & 0.3529 & 0.2974 & 0.1308 & 0.2641 \\
    & OWLv2      & 0.3117 & 0.2474 & 0.0898 & 0.2194 \\
\midrule
Qwen2.5-VL-7B
    & YOLO-World & 0.4067 & 0.3356 & 0.1471 & 0.3004 \\
    & G-DINO-B   & 0.3700 & 0.3098 & 0.1292 & 0.2737 \\
    & G-DINO-T   & 0.3467 & 0.2908 & 0.1303 & 0.2594 \\
    & OWLv2      & 0.3163 & 0.2509 & 0.0917 & 0.2227 \\
\midrule
Gemma4-E2B
    & YOLO-World & 0.3651 & 0.3057 & 0.1441 & 0.2750 \\
    & G-DINO-B   & 0.3379 & 0.2899 & 0.1342 & 0.2576 \\
    & G-DINO-T   & 0.3379 & 0.2907 & 0.1431 & 0.2606 \\
    & OWLv2      & 0.2819 & 0.2287 & 0.0891 & 0.2028 \\
\midrule
Gemma3-4B
    & YOLO-World & 0.3862 & 0.3181 & 0.1392 & 0.2849 \\
    & G-DINO-B   & 0.3445 & 0.2878 & 0.1199 & 0.2545 \\
    & G-DINO-T   & 0.3186 & 0.2664 & 0.1217 & 0.2386 \\
    & OWLv2      & 0.3031 & 0.2403 & 0.0883 & 0.2135 \\
\midrule
InternVL3-8B
    & YOLO-World & 0.3818 & 0.3144 & 0.1371 & 0.2815 \\
    & G-DINO-B   & 0.3497 & 0.2914 & 0.1197 & 0.2574 \\
    & G-DINO-T   & 0.3170 & 0.2642 & 0.1172 & 0.2360 \\
    & OWLv2      & 0.3044 & 0.2408 & 0.0881 & 0.2141 \\
\midrule
Gemma4-12B
    & YOLO-World & 0.3697 & 0.3040 & 0.1330 & 0.2724 \\
    & G-DINO-B   & 0.3249 & 0.2683 & 0.1119 & 0.2384 \\
    & G-DINO-T   & 0.2997 & 0.2483 & 0.1102 & 0.2223 \\
    & OWLv2      & 0.2808 & 0.2209 & 0.0813 & 0.1970 \\
\bottomrule
\end{tabular}
\end{table*}

\begin{table*}[t]
\centering
\caption{Threshold-wise SNAP on COCO-Val using Sentence-BERT for semantic
similarity. LO, MI, and HI correspond to similarity thresholds
$\{0.50,0.55,0.60\}$, $\{0.65,0.70,0.75,0.80\}$, and
$\{0.85,0.90,0.95\}$, respectively. Macro is averaged over all ten
thresholds.}
\label{tab:app_coco_snap_intervals}
\scriptsize
\setlength{\tabcolsep}{5pt}
\renewcommand{\arraystretch}{0.93}
\begin{tabular}{llcccc}
\toprule
MLLM & Detector & LO & MI & HI & Macro \\
\midrule
Gemma4-E4B
    & YOLO-World & 0.3960 & 0.2927 & 0.2289 & 0.3045 \\
    & G-DINO-B   & 0.3158 & 0.2079 & 0.1541 & 0.2241 \\
    & G-DINO-T   & 0.3292 & 0.2299 & 0.1762 & 0.2436 \\
    & OWLv2      & 0.2879 & 0.2226 & 0.1579 & 0.2228 \\
\midrule
Qwen3.5-9B
    & YOLO-World & 0.4191 & 0.2989 & 0.2191 & 0.3110 \\
    & G-DINO-B   & 0.3255 & 0.2094 & 0.1492 & 0.2262 \\
    & G-DINO-T   & 0.3258 & 0.2195 & 0.1577 & 0.2328 \\
    & OWLv2      & 0.2966 & 0.2261 & 0.1501 & 0.2245 \\
\midrule
Qwen2.5-VL-7B
    & YOLO-World & 0.4314 & 0.3016 & 0.2031 & 0.3110 \\
    & G-DINO-B   & 0.3319 & 0.2084 & 0.1347 & 0.2234 \\
    & G-DINO-T   & 0.3306 & 0.2227 & 0.1480 & 0.2327 \\
    & OWLv2      & 0.3052 & 0.2345 & 0.1444 & 0.2287 \\
\midrule
Gemma4-E2B
    & YOLO-World & 0.3405 & 0.2494 & 0.1950 & 0.2604 \\
    & G-DINO-B   & 0.2750 & 0.1803 & 0.1362 & 0.1955 \\
    & G-DINO-T   & 0.2885 & 0.2000 & 0.1551 & 0.2131 \\
    & OWLv2      & 0.2554 & 0.1950 & 0.1396 & 0.1965 \\
\midrule
Gemma3-4B
    & YOLO-World & 0.3927 & 0.2554 & 0.1660 & 0.2698 \\
    & G-DINO-B   & 0.2934 & 0.1725 & 0.1129 & 0.1909 \\
    & G-DINO-T   & 0.2839 & 0.1770 & 0.1151 & 0.1905 \\
    & OWLv2      & 0.2840 & 0.2169 & 0.1349 & 0.2124 \\
\midrule
InternVL3-8B
    & YOLO-World & 0.4044 & 0.2754 & 0.1735 & 0.2835 \\
    & G-DINO-B   & 0.3160 & 0.1951 & 0.1207 & 0.2091 \\
    & G-DINO-T   & 0.3044 & 0.1975 & 0.1227 & 0.2071 \\
    & OWLv2      & 0.2980 & 0.2289 & 0.1387 & 0.2226 \\
\midrule
Gemma4-12B
    & YOLO-World & 0.3838 & 0.2835 & 0.2016 & 0.2890 \\
    & G-DINO-B   & 0.2788 & 0.1879 & 0.1326 & 0.1986 \\
    & G-DINO-T   & 0.2783 & 0.1995 & 0.1403 & 0.2054 \\
    & OWLv2      & 0.2663 & 0.2073 & 0.1360 & 0.2036 \\
\bottomrule
\end{tabular}
\end{table*}

\begin{table*}[t]
\centering
\caption{SemLoc-AP threshold marginals on COCO-Val, averaged over all 28
MLLM--detector combinations. The first row fixes the IoU threshold $\theta$
and averages over all ten semantic thresholds $\tau$; the second fixes
$\tau$ and averages over all ten IoU thresholds $\theta$. The overall
SemLoc-AP averaged over all 100 threshold pairs is $0.1410$.}
\label{tab:app_coco_semloc_marginals}
\scriptsize
\setlength{\tabcolsep}{3.5pt}
\renewcommand{\arraystretch}{1.05}
\resizebox{\textwidth}{!}{
\begin{tabular}{lcccccccccc}
\toprule
 & 0.50 & 0.55 & 0.60 & 0.65 & 0.70 & 0.75 & 0.80 & 0.85 & 0.90 & 0.95 \\
\midrule
$S_{\mathrm{IoU}}(\theta)$
& 0.1906 & 0.1861 & 0.1810 & 0.1748 & 0.1667
& 0.1547 & 0.1383 & 0.1140 & 0.0780 & 0.0257 \\

$S_{\mathrm{sem}}(\tau)$
& 0.1875 & 0.1802 & 0.1758 & 0.1483 & 0.1415
& 0.1358 & 0.1300 & 0.1194 & 0.1000 & 0.0913 \\
\bottomrule
\end{tabular}
}
\end{table*}

The SemLoc-AP marginals show that joint correctness decreases as either
criterion becomes more stringent. On COCO-Val, increasing the IoU threshold
produces the larger reduction, particularly at the strictest thresholds,
indicating that precise localization is a stronger limiting factor for joint
semantic--spatial correctness in these experiments.

\subsection{LVIS-minival}

Table~\ref{tab:app_lvis_intervals} provides the corresponding threshold-wise
CAAP and SNAP results on LVIS-minival. The detector ranking differs from
COCO-Val across the threshold ranges: OWLv2 consistently provides the strongest
CAAP and SNAP performance for the evaluated MLLMs, further illustrating the
dependence of the grounding stage on the target category distribution.
Table~\ref{tab:app_lvis_semloc_marginals} additionally reports the SemLoc-AP
threshold marginals averaged across all 12 MLLM--detector combinations.

\begin{table*}[t]
\centering
\caption{Threshold-wise CAAP and SNAP on LVIS-minival. Macro scores are
averaged over all ten thresholds rather than over the three interval means.}
\label{tab:app_lvis_intervals}
\scriptsize
\setlength{\tabcolsep}{3.6pt}
\renewcommand{\arraystretch}{0.95}
\begin{tabular}{llcccccccc}
\toprule
& & \multicolumn{4}{c}{CAAP} & \multicolumn{4}{c}{SNAP} \\
\cmidrule(lr){3-6}
\cmidrule(lr){7-10}
MLLM & Detector
& LO & MI & HI & Macro
& LO & MI & HI & Macro \\
\midrule

Qwen3.5-9B
& YOLO-World & 0.2306 & 0.1867 & 0.0944 & 0.1722
             & 0.2628 & 0.1875 & 0.0993 & 0.1836 \\
& G-DINO-B   & 0.1777 & 0.1436 & 0.0632 & 0.1297
             & 0.2026 & 0.1431 & 0.0793 & 0.1418 \\
& G-DINO-T   & 0.2025 & 0.1669 & 0.0833 & 0.1525
             & 0.2248 & 0.1575 & 0.0886 & 0.1570 \\
& OWLv2      & 0.2908 & 0.2294 & 0.0915 & 0.2064
             & 0.3276 & 0.2427 & 0.1326 & 0.2352 \\

\midrule

Gemma4-12B
& YOLO-World & 0.2351 & 0.1904 & 0.0948 & 0.1751
             & 0.2623 & 0.1827 & 0.0899 & 0.1787 \\
& G-DINO-B   & 0.1725 & 0.1396 & 0.0615 & 0.1260
             & 0.1939 & 0.1320 & 0.0684 & 0.1315 \\
& G-DINO-T   & 0.1992 & 0.1632 & 0.0797 & 0.1489
             & 0.2130 & 0.1446 & 0.0755 & 0.1444 \\
& OWLv2      & 0.2926 & 0.2307 & 0.0903 & 0.2071
             & 0.3210 & 0.2324 & 0.1172 & 0.2244 \\

\midrule

Gemma4-E4B
& YOLO-World & 0.1961 & 0.1575 & 0.0811 & 0.1462
             & 0.2081 & 0.1443 & 0.0788 & 0.1438 \\
& G-DINO-B   & 0.1497 & 0.1192 & 0.0540 & 0.1088
             & 0.1604 & 0.1066 & 0.0587 & 0.1084 \\
& G-DINO-T   & 0.1767 & 0.1450 & 0.0767 & 0.1340
             & 0.1789 & 0.1189 & 0.0657 & 0.1209 \\
& OWLv2      & 0.2440 & 0.1942 & 0.0813 & 0.1753
             & 0.2655 & 0.1909 & 0.1072 & 0.1881 \\

\bottomrule
\end{tabular}
\end{table*}

\begin{table*}[t]
\centering
\caption{SemLoc-AP threshold marginals on LVIS-minival, averaged over all 12
MLLM--detector combinations. The first row fixes the IoU threshold $\theta$
and averages over all ten semantic thresholds $\tau$; the second fixes
$\tau$ and averages over all ten IoU thresholds $\theta$. The overall
SemLoc-AP averaged over all 100 threshold pairs is $0.0893$.}
\label{tab:app_lvis_semloc_marginals}
\scriptsize
\setlength{\tabcolsep}{3.5pt}
\renewcommand{\arraystretch}{1.05}
\resizebox{\textwidth}{!}{
\begin{tabular}{lcccccccccc}
\toprule
 & 0.50 & 0.55 & 0.60 & 0.65 & 0.70 & 0.75 & 0.80 & 0.85 & 0.90 & 0.95 \\
\midrule
$S_{\mathrm{IoU}}(\theta)$
& 0.1210 & 0.1177 & 0.1138 & 0.1092 & 0.1034
& 0.0959 & 0.0859 & 0.0715 & 0.0513 & 0.0230 \\

$S_{\mathrm{sem}}(\tau)$
& 0.1231 & 0.1175 & 0.1133 & 0.1061 & 0.1005
& 0.0874 & 0.0774 & 0.0707 & 0.0535 & 0.0433 \\
\bottomrule
\end{tabular}
}
\end{table*}

As on COCO-Val, SemLoc-AP decreases as either spatial or semantic matching
becomes more stringent. The marginals also show a larger reduction at the
strictest IoU thresholds than at the strictest semantic thresholds, indicating
that precise localization remains an important limiting factor for joint
semantic--spatial correctness on LVIS.

\subsection{COCO-OOD}

Table~\ref{tab:app_ood_caap_intervals} reports the complete CAAP breakdown for
COCO-OOD. SNAP and SemLoc-AP are not reported on this dataset because all
annotated out-of-distribution objects share a single generic unknown category,
so the benchmark does not provide meaningful semantic labels against which
generated object names or joint semantic--spatial correctness can be
evaluated.

\begin{table*}[t]
\centering
\caption{Threshold-wise CAAP on COCO-OOD for all 28 MLLM--detector
combinations. Macro is averaged over all ten IoU thresholds from $0.50$ to
$0.95$.}
\label{tab:app_ood_caap_intervals}
\scriptsize
\setlength{\tabcolsep}{5pt}
\renewcommand{\arraystretch}{0.93}
\begin{tabular}{llcccc}
\toprule
MLLM & Detector & LO & MI & HI & Macro \\
\midrule

Gemma4-E4B
& YOLO-World & 0.7503 & 0.6873 & 0.4267 & 0.6280 \\
& G-DINO-B   & 0.6208 & 0.5739 & 0.3430 & 0.5187 \\
& G-DINO-T   & 0.6624 & 0.6134 & 0.3839 & 0.5592 \\
& OWLv2      & 0.6945 & 0.6319 & 0.3391 & 0.5628 \\

\midrule

Qwen2.5-VL-7B
& YOLO-World & 0.7611 & 0.6953 & 0.4310 & 0.6357 \\
& G-DINO-B   & 0.5854 & 0.5431 & 0.3241 & 0.4901 \\
& G-DINO-T   & 0.6190 & 0.5659 & 0.3528 & 0.5179 \\
& OWLv2      & 0.7011 & 0.6354 & 0.3377 & 0.5658 \\

\midrule

InternVL3-8B
& YOLO-World & 0.7655 & 0.7004 & 0.4351 & 0.6403 \\
& G-DINO-B   & 0.5861 & 0.5418 & 0.3191 & 0.4883 \\
& G-DINO-T   & 0.6050 & 0.5528 & 0.3355 & 0.5032 \\
& OWLv2      & 0.7035 & 0.6382 & 0.3396 & 0.5682 \\

\midrule

Qwen3.5-9B
& YOLO-World & 0.7570 & 0.6902 & 0.4249 & 0.6306 \\
& G-DINO-B   & 0.5856 & 0.5407 & 0.3194 & 0.4878 \\
& G-DINO-T   & 0.6204 & 0.5626 & 0.3451 & 0.5147 \\
& OWLv2      & 0.6984 & 0.6329 & 0.3367 & 0.5637 \\

\midrule

Gemma4-E2B
& YOLO-World & 0.7118 & 0.6533 & 0.4163 & 0.5998 \\
& G-DINO-B   & 0.5945 & 0.5574 & 0.3499 & 0.5063 \\
& G-DINO-T   & 0.6331 & 0.5890 & 0.3863 & 0.5414 \\
& OWLv2      & 0.6493 & 0.5907 & 0.3190 & 0.5268 \\

\midrule

Gemma3-4B
& YOLO-World & 0.7470 & 0.6817 & 0.4233 & 0.6238 \\
& G-DINO-B   & 0.5669 & 0.5274 & 0.3196 & 0.4769 \\
& G-DINO-T   & 0.5995 & 0.5471 & 0.3386 & 0.5003 \\
& OWLv2      & 0.6841 & 0.6193 & 0.3305 & 0.5521 \\

\midrule

Gemma4-12B
& YOLO-World & 0.7490 & 0.6881 & 0.4303 & 0.6290 \\
& G-DINO-B   & 0.5717 & 0.5290 & 0.3096 & 0.4760 \\
& G-DINO-T   & 0.5931 & 0.5413 & 0.3265 & 0.4924 \\
& OWLv2      & 0.6785 & 0.6161 & 0.3268 & 0.5480 \\

\bottomrule
\end{tabular}
\end{table*}

\section{Complete COCO-OOD Unknown-Object Results}
\label{app:ood_full}

The main paper reports the Qwen3.5-9B + YOLO-World configuration in the
comparison against existing unknown-object detection methods. This configuration
is used as our main model because it is selected from the COCO-Val experiments,
rather than chosen specifically to maximize COCO-OOD performance. For
completeness, Table~\ref{tab:app_ood_full} reports unknown-object detection
results for all 28 combinations of seven MLLMs and four open-vocabulary
detectors.

We report Unknown Average Precision (U-AP), together with the maximum Unknown
F1 score (U-F1) obtained by sweeping the detection scores for each
configuration. U-PRE and U-REC denote the precision and recall at the same
operating point at which U-F1 is maximized. This reporting sweep is distinct
from the fixed detector-side confidence threshold used to generate predictions,
which is selected on held-out data and kept fixed for each detector. The
U-F1 operating point is instead selected from each configuration's score
ranking, avoiding the use of a shared numerical score cutoff across detectors
with different score distributions.

\begin{table*}[t]
\centering
\caption{Complete unknown-object detection results on COCO-OOD for all
28 MLLM--detector combinations. U-PRE and U-REC are measured at the operating
point that maximizes U-F1 for each configuration. $\dagger$ denotes the main
Qwen3.5-9B + YOLO-World configuration reported in the main-paper comparison.}
\label{tab:app_ood_full}
\scriptsize
\setlength{\tabcolsep}{6pt}
\renewcommand{\arraystretch}{0.95}
\begin{tabular}{llcccc}
\toprule
MLLM & Detector & U-AP & U-F1 & U-PRE & U-REC \\
\midrule

Gemma4-E4B
& YOLO-World & 0.762 & \textbf{0.721} & \textbf{0.804} & 0.653 \\
& G-DINO-B   & 0.629 & 0.642 & 0.626 & 0.658 \\
& G-DINO-T   & 0.671 & 0.666 & 0.634 & 0.701 \\
& OWLv2      & 0.703 & 0.666 & 0.745 & 0.602 \\

\midrule

Qwen3.5-9B
& YOLO-World$^\dagger$ & 0.769 & 0.720 & 0.787 & 0.663 \\
& G-DINO-B             & 0.591 & 0.593 & 0.519 & 0.692 \\
& G-DINO-T             & 0.629 & 0.612 & 0.561 & 0.675 \\
& OWLv2                & 0.707 & 0.661 & 0.699 & 0.627 \\

\midrule

Qwen2.5-VL-7B
& YOLO-World & 0.772 & 0.720 & 0.759 & 0.685 \\
& G-DINO-B   & 0.592 & 0.594 & 0.530 & 0.677 \\
& G-DINO-T   & 0.628 & 0.615 & 0.570 & 0.668 \\
& OWLv2      & 0.709 & 0.659 & 0.634 & 0.688 \\

\midrule

Gemma4-E2B
& YOLO-World & 0.723 & 0.700 & 0.780 & 0.636 \\
& G-DINO-B   & 0.599 & 0.613 & 0.586 & 0.642 \\
& G-DINO-T   & 0.641 & 0.643 & 0.672 & 0.615 \\
& OWLv2      & 0.656 & 0.636 & 0.714 & 0.574 \\

\midrule

Gemma3-4B
& YOLO-World & 0.760 & 0.707 & 0.761 & 0.659 \\
& G-DINO-B   & 0.574 & 0.583 & 0.547 & 0.624 \\
& G-DINO-T   & 0.608 & 0.599 & 0.578 & 0.622 \\
& OWLv2      & 0.692 & 0.651 & 0.643 & 0.659 \\

\midrule

InternVL3-8B
& YOLO-World & \textbf{0.776} & 0.720 & 0.783 & 0.666 \\
& G-DINO-B   & 0.593 & 0.594 & 0.507 & \textbf{0.717} \\
& G-DINO-T   & 0.612 & 0.602 & 0.535 & 0.688 \\
& OWLv2      & 0.712 & 0.656 & 0.696 & 0.621 \\

\midrule

Gemma4-12B
& YOLO-World & 0.760 & 0.701 & 0.803 & 0.622 \\
& G-DINO-B   & 0.578 & 0.577 & 0.508 & 0.667 \\
& G-DINO-T   & 0.601 & 0.579 & 0.508 & 0.672 \\
& OWLv2      & 0.685 & 0.635 & 0.595 & 0.680 \\

\bottomrule
\end{tabular}
\end{table*}

The complete results show that the strong COCO-OOD performance reported in the
main paper is not specific to a single MLLM--detector pairing. YOLO-World
provides the strongest U-AP and U-F1 performance for every evaluated MLLM, with
U-AP ranging from $0.723$ to $0.776$. The main Qwen3.5-9B + YOLO-World
configuration achieves U-AP $0.769$ and U-F1 $0.720$, close to the strongest
values observed anywhere in the full grid.

The highest U-AP is obtained by InternVL3-8B + YOLO-World at $0.776$, while
Gemma4-E4B + YOLO-World achieves the highest U-F1 at $0.721$ and the highest
U-PRE at $0.804$. The highest U-REC, $0.717$, is obtained by InternVL3-8B +
Grounding DINO-B. These differences also illustrate the precision--recall
trade-offs among grounding models: YOLO-World generally provides the strongest
overall balance, whereas some Grounding DINO configurations obtain higher
recall at the cost of substantially lower precision and U-AP.

\section{Detailed Prompt-Sensitivity Results}
\label{app:prompt_ablation}

The main paper summarizes the prompt-sensitivity experiment using the three
MLLMs evaluated on both COCO-Val and LVIS-minival. Here, we provide the full
per-MLLM CAAP and SNAP results for each dataset, together with the generated
vocabulary sizes. The experiment uses frozen 500-image subsets from the
corresponding test splits, with YOLO-World as the grounding model at the fixed
confidence threshold of $0.001$. The COCO-Val subset contains 3,903
ground-truth objects and the LVIS-minival subset contains 4,294. Both subsets
are disjoint from the hyperparameter-selection holdout.

We evaluate three prompt conditions. \emph{Default} is the generic prompt used
for the main experiments; \emph{Minimal} removes the additional output and
category constraints and simply asks the MLLM to list the objects in the
image; and \emph{COCO-specific} adds constraints encouraging general category
names, excluding infrastructural objects, and requesting singular object
names. The same generated vocabularies are used to evaluate CAAP, SNAP, and
SemLoc-AP.

\subsection{COCO-Val}

Tables~\ref{tab:app_prompt_coco_caap} and
\ref{tab:app_prompt_coco_snap} report the complete COCO-Val results across all
seven MLLMs. The Default and Minimal prompts produce very similar CAAP values:
averaged across all seven models, CAAP is $0.2828$ for Default and $0.2834$
for Minimal. In contrast, the COCO-specific prompt increases the mean CAAP to
$0.3091$ and improves CAAP for every evaluated MLLM.

\begin{table}[t]
\centering
\caption{CAAP on the 500-image COCO-Val prompt-ablation subset.}
\label{tab:app_prompt_coco_caap}
\scriptsize
\setlength{\tabcolsep}{4.5pt}
\renewcommand{\arraystretch}{0.95}
\begin{tabular}{lccc}
\toprule
MLLM & Default & Minimal & COCO-specific \\
\midrule
Gemma3-4B       & 0.2813 & 0.2871 & 0.2894 \\
Gemma4-12B      & 0.2699 & 0.2799 & \textbf{0.3417} \\
Gemma4-E2B      & 0.2680 & 0.2652 & 0.2886 \\
Gemma4-E4B      & 0.2928 & 0.2936 & 0.3224 \\
InternVL3-8B    & 0.2785 & 0.2764 & 0.2894 \\
Qwen2.5-VL-7B   & 0.2943 & 0.2990 & 0.3190 \\
Qwen3.5-9B      & 0.2945 & 0.2829 & 0.3129 \\
\midrule
Mean            & 0.2828 & 0.2834 & \textbf{0.3091} \\
\bottomrule
\end{tabular}
\end{table}

The effect is even more pronounced under SNAP. As shown in
Table~\ref{tab:app_prompt_coco_snap}, the COCO-specific prompt raises the
seven-model mean SNAP from $0.2842$ under Default to $0.3648$. The largest
increase occurs for Gemma4-12B, for which SNAP increases from $0.2923$ to
$0.4681$.

\begin{table}[t]
\centering
\caption{SNAP on the 500-image COCO-Val prompt-ablation subset. Semantic
similarity is computed using Sentence-BERT.}
\label{tab:app_prompt_coco_snap}
\scriptsize
\setlength{\tabcolsep}{4.5pt}
\renewcommand{\arraystretch}{0.95}
\begin{tabular}{lccc}
\toprule
MLLM & Default & Minimal & COCO-specific \\
\midrule
Gemma3-4B       & 0.2630 & 0.2647 & 0.2897 \\
Gemma4-12B      & 0.2923 & 0.2892 & \textbf{0.4681} \\
Gemma4-E2B      & 0.2596 & 0.2561 & 0.3308 \\
Gemma4-E4B      & 0.2955 & 0.2897 & 0.3989 \\
InternVL3-8B    & 0.2771 & 0.2696 & 0.3086 \\
Qwen2.5-VL-7B   & 0.3031 & 0.3081 & 0.3796 \\
Qwen3.5-9B      & 0.2986 & 0.2771 & 0.3779 \\
\midrule
Mean            & 0.2842 & 0.2792 & \textbf{0.3648} \\
\bottomrule
\end{tabular}
\end{table}

Table~\ref{tab:app_prompt_coco_labels} shows the corresponding number of
labels proposed per image. The COCO-specific prompt reduces the average
vocabulary size from $7.6$ labels per image under Default to $6.2$. Thus, its
higher COCO performance does not arise from proposing more candidate concepts.
Instead, the prompt suppresses categories that are less aligned with the
relatively coarse COCO annotation vocabulary.

\begin{table}[t]
\centering
\caption{Average number of labels generated per image on the COCO-Val
prompt-ablation subset.}
\label{tab:app_prompt_coco_labels}
\scriptsize
\setlength{\tabcolsep}{4.5pt}
\renewcommand{\arraystretch}{0.95}
\begin{tabular}{lccc}
\toprule
MLLM & Default & Minimal & COCO-specific \\
\midrule
Gemma3-4B       & 8.0  & 8.1  & 6.8 \\
Gemma4-12B      & 11.8 & 17.6 & 9.3 \\
Gemma4-E2B      & 3.6  & 3.8  & 2.8 \\
Gemma4-E4B      & 4.3  & 4.7  & 3.6 \\
InternVL3-8B    & 8.6  & 9.0  & 7.1 \\
Qwen2.5-VL-7B   & 8.5  & 8.2  & 6.9 \\
Qwen3.5-9B      & 8.2  & 9.6  & 7.2 \\
\midrule
Mean            & 7.6  & 8.7  & 6.2 \\
\bottomrule
\end{tabular}
\end{table}

\subsection{LVIS-minival}

The same prompt variants behave differently on LVIS-minival. As shown in
Tables~\ref{tab:app_prompt_lvis_caap} and
\ref{tab:app_prompt_lvis_snap}, the COCO-specific prompt reduces both CAAP and
SNAP for all three evaluated MLLMs. Mean CAAP decreases from $0.1737$ under
Default to $0.1514$, while mean SNAP decreases from $0.1821$ to $0.1651$.

\begin{table}[t]
\centering
\caption{CAAP on the 500-image LVIS-minival prompt-ablation subset.}
\label{tab:app_prompt_lvis_caap}
\scriptsize
\setlength{\tabcolsep}{4.5pt}
\renewcommand{\arraystretch}{0.95}
\begin{tabular}{lccc}
\toprule
MLLM & Default & Minimal & COCO-specific \\
\midrule
Gemma4-12B      & \textbf{0.1855} & 0.1840 & 0.1597 \\
Gemma4-E4B      & \textbf{0.1548} & 0.1487 & 0.1264 \\
Qwen3.5-9B      & 0.1808 & \textbf{0.1828} & 0.1681 \\
\midrule
Mean            & \textbf{0.1737} & 0.1718 & 0.1514 \\
\bottomrule
\end{tabular}
\end{table}

\begin{table}[t]
\centering
\caption{SNAP on the 500-image LVIS-minival prompt-ablation subset. Semantic
similarity is computed using Sentence-BERT.}
\label{tab:app_prompt_lvis_snap}
\scriptsize
\setlength{\tabcolsep}{4.5pt}
\renewcommand{\arraystretch}{0.95}
\begin{tabular}{lccc}
\toprule
MLLM & Default & Minimal & COCO-specific \\
\midrule
Gemma4-12B      & \textbf{0.1933} & 0.1863 & 0.1761 \\
Gemma4-E4B      & \textbf{0.1564} & 0.1482 & 0.1279 \\
Qwen3.5-9B      & \textbf{0.1965} & 0.1891 & 0.1913 \\
\midrule
Mean            & \textbf{0.1821} & 0.1745 & 0.1651 \\
\bottomrule
\end{tabular}
\end{table}

\begin{table}[t]
\centering
\caption{Average number of labels generated per image on the LVIS-minival
prompt-ablation subset.}
\label{tab:app_prompt_lvis_labels}
\scriptsize
\setlength{\tabcolsep}{4.5pt}
\renewcommand{\arraystretch}{0.95}
\begin{tabular}{lccc}
\toprule
MLLM & Default & Minimal & COCO-specific \\
\midrule
Gemma4-12B      & 11.7 & 17.3 & 9.0 \\
Gemma4-E4B      & 4.2  & 4.6  & 3.5 \\
Qwen3.5-9B      & 7.8  & 9.5  & 6.7 \\
\midrule
Mean            & 7.9  & 10.5 & 6.4 \\
\bottomrule
\end{tabular}
\end{table}

For the three MLLMs common to both datasets, the joint SemLoc-AP results follow
the same cross-dataset trend summarized in the main paper. On COCO-Val,
SemLoc-AP increases from $0.1740$ under Default to $0.2204$ under the
COCO-specific prompt, while the Minimal prompt yields $0.1707$. On
LVIS-minival, Default performs best at $0.1058$, followed by Minimal at
$0.1026$ and COCO-specific at $0.0983$. Thus, the prompt specialization that
improves localization and semantic naming on COCO also improves their correct
association, but this advantage does not transfer to the finer-grained LVIS
setting.

\subsection{Effect of Dataset-Specific Prompt Constraints}

The opposite behavior on COCO and LVIS can be traced to the assumptions
encoded by the COCO-specific prompt. First, asking for only the
\emph{general name} of each object agrees well with the relatively coarse COCO
ontology, but conflicts with the substantially more fine-grained LVIS
vocabulary, which contains categories such as \emph{Tabasco sauce},
\emph{beer bottle}, and \emph{fish (food)}. Second, suppressing
\emph{infrastructural objects} has little cost under COCO but can remove valid
LVIS categories such as \emph{fireplug}, \emph{streetlight},
\emph{telephone pole}, and \emph{street sign}. Finally, enforcing singular
category names agrees with COCO's naming convention, whereas LVIS contains
categories whose canonical names are plural, including \emph{asparagus},
\emph{binoculars}, and \emph{suspenders}.

The vocabulary-size results support this interpretation. The COCO-specific
prompt produces fewer labels per image on both datasets. On COCO, this
suppression primarily removes concepts that are not represented by the
benchmark ontology and consequently improves CAAP and SNAP. On LVIS, the same
constraints suppress valid fine-grained categories and reduce performance.
The corresponding SemLoc-AP results follow the same pattern, indicating that
these prompt constraints also affect whether the generated semantic names are
correctly associated with the localized objects. Thus, the gain from the
specialized prompt primarily reflects better agreement with a particular
annotation convention rather than a generally improved ability to recognize
objects.

For the three MLLMs evaluated on both datasets, the COCO-specific prompt
improves CAAP over Default by an average of $14.3\%$ on COCO-Val, but decreases
it by $13.1\%$ on LVIS-minival. These percentages are computed as the mean of
the per-MLLM relative changes, rather than from the ratio of the aggregate mean
scores. In contrast, changing only the generic wording from Default to Minimal
has little effect: the mean CAAP change is $+0.3\%$ across the seven COCO
models and $-1.2\%$ across the three LVIS models. These results motivate our
use of a single generic Default prompt across all datasets rather than
selecting a separately optimized prompt for each benchmark.

\section{Closing the MLLM--OVOD Loop: Selective Refinement Analysis}
\label{app:selective_refinement}

The main \LAOD{} formulation is intentionally one-pass: the MLLM interprets
the image and constructs an image-specific detection vocabulary, after which
the open-vocabulary detector grounds the generated concepts. A natural
extension is to close this loop by returning information from the detector to
the MLLM, allowing the vocabulary to be reconsidered before a second grounding
pass. We therefore investigate whether such MLLM--OVOD interaction provides a
meaningful advantage over the simpler one-pass formulation.

Let $\mathcal{D}^{(0)}$ denote the detections produced by standard \LAOD{}.
To close the loop, detector feedback from the first pass is provided to the
MLLM, which generates a revised vocabulary. The same open-vocabulary detector
is then applied again to obtain a refined detection set
$\mathcal{D}^{(1)}$. We consider two refinement strategies. Under
\emph{Always Refine}, this additional MLLM interaction and detector pass are
performed for every image. Under \emph{Selective Refinement}, the second pass
is invoked only for images whose first-pass detections have relatively low
detector confidence. If refinement is not invoked, the original
$\mathcal{D}^{(0)}$ is retained; otherwise, $\mathcal{D}^{(1)}$ is used
directly. The two detection sets are never merged.

We evaluate this behavior retrospectively on the frozen 500-image COCO-Val
analysis subset using three cached MLLM--detector configurations:
Qwen3.5-9B with YOLO-World, Gemma4-E4B with YOLO-World, and Qwen3.5-9B
with OWLv2. For all three configurations, the first- and second-pass outputs
cover identical image sets and use identical detector settings between rounds.
Because both passes were already cached, this analysis requires no additional
model inference; however, we report the nominal inference requirements that
would be incurred when deploying each strategy.

\paragraph{Selecting a common refinement budget.}
For selective refinement, images are ranked according to the maximum confidence
of their first-pass detections,
\begin{equation}
s_{\max}(X)
=
\max_i s_i,
\end{equation}
and refinement is applied to the lower-confidence fraction of the dataset.
We first examined the refinement rate using Qwen3.5-9B with YOLO-World.
Performance improved smoothly from low refinement rates and reached a broad
high-performing region between approximately $55\%$ and $75\%$, with CAAP,
SNAP, and SemLoc-AP attaining their best or near-best values around
$60$--$65\%$. We therefore use a fixed $60\%$ refinement budget for the
controlled cross-configuration comparison below. This operating point is
intended to represent the stable region observed in the initial analysis rather
than to imply that $60\%$ is universally optimal. In particular, no
configuration-specific refinement rate is selected for the other model pairs.

Because detector confidence values are not calibrated across YOLO-World and
OWLv2, the primary comparison uses the lowest-confidence $60\%$ of images
within each configuration rather than transferring an absolute score threshold
between detector families. At this refinement rate, the system requires on
average $1.6$ MLLM calls and $1.6$ OVOD passes per image, compared with one
of each for standard \LAOD{} and two of each for unconditional refinement.

\begin{table*}[t]
\centering
\caption{Comparison of standard one-pass \LAOD{}, unconditional refinement,
and selective MLLM--OVOD refinement on the 500-image COCO-Val analysis
subset. Selective refinement applies one additional MLLM interaction and OVOD
pass to the $60\%$ of images with the lowest first-pass maximum detector
confidence. The same refinement budget is used for all configurations.
MLLM Calls/Img and OVOD Passes/Img denote the nominal inference requirements;
the outputs used in this retrospective analysis were already cached. Bold
values indicate the best result within each configuration and metric.}
\label{tab:selective_refinement}
\small
\setlength{\tabcolsep}{4pt}
\renewcommand{\arraystretch}{1.05}
\resizebox{\textwidth}{!}{
\begin{tabular}{llcccccc}
\toprule
Configuration & Method &
Refined (\%) &
MLLM Calls/Img &
OVOD Passes/Img &
CAAP & SNAP & SemLoc-AP \\
\midrule

\multirow{3}{*}{Qwen3.5-9B + YOLO-World}
& Standard \LAOD{}
& 0 & 1.00 & 1.00
& 0.2809 & 0.2957 & 0.1712 \\
& Always Refine
& 100 & 2.00 & 2.00
& 0.2807 & 0.3030 & 0.1730 \\
& Selective Refinement
& 60 & 1.60 & 1.60
& \textbf{0.2844} & \textbf{0.3076} & \textbf{0.1759} \\

\midrule

\multirow{3}{*}{Gemma4-E4B + YOLO-World}
& Standard \LAOD{}
& 0 & 1.00 & 1.00
& \textbf{0.2780} & 0.2898 & 0.1741 \\
& Always Refine
& 100 & 2.00 & 2.00
& 0.2740 & 0.2910 & 0.1760 \\
& Selective Refinement
& 60 & 1.60 & 1.60
& 0.2777 & \textbf{0.2945} & \textbf{0.1782} \\

\midrule

\multirow{3}{*}{Qwen3.5-9B + OWLv2}
& Standard \LAOD{}
& 0 & 1.00 & 1.00
& \textbf{0.2086} & \textbf{0.2198} & \textbf{0.1199} \\
& Always Refine
& 100 & 2.00 & 2.00
& 0.2068 & 0.2114 & 0.1150 \\
& Selective Refinement
& 60 & 1.60 & 1.60
& 0.2080 & 0.2127 & 0.1160 \\

\bottomrule
\end{tabular}
}
\end{table*}

Table~\ref{tab:selective_refinement} shows that closing the MLLM--OVOD loop
does not provide a consistent improvement across configurations. For
Qwen3.5-9B with YOLO-World, selective refinement improves all three metrics,
increasing CAAP from $0.2809$ to $0.2844$, SNAP from $0.2957$ to $0.3076$,
and SemLoc-AP from $0.1712$ to $0.1759$. These correspond to relative gains
of approximately $1.3\%$, $4.0\%$, and $2.7\%$, respectively. Unconditional
refinement is less effective: it slightly reduces CAAP to $0.2807$ while
increasing SNAP and SemLoc-AP only to $0.3030$ and $0.1730$.

The behavior is already less consistent when the MLLM is changed while keeping
the grounding model fixed. With Gemma4-E4B and YOLO-World, selective
refinement improves SNAP from $0.2898$ to $0.2945$ and SemLoc-AP from
$0.1741$ to $0.1782$, but CAAP slightly decreases from $0.2780$ to
$0.2777$. Unconditional refinement reduces CAAP further to $0.2740$. Thus,
the semantic gains obtained by closing the loop do not necessarily translate
into improved localization.

The result changes more substantially when the grounding detector is replaced.
For Qwen3.5-9B with OWLv2, both refinement policies are detrimental.
Selective refinement decreases CAAP from $0.2086$ to $0.2080$, SNAP from
$0.2198$ to $0.2127$, and SemLoc-AP from $0.1199$ to $0.1160$.
Unconditional refinement reduces the corresponding scores further to
$0.2068$, $0.2114$, and $0.1150$. Overall, only one of the three evaluated
configurations improves across all three metrics, and the effectiveness of the
additional MLLM--OVOD interaction is therefore strongly configuration
dependent.

\begin{figure*}[t]
    \centering
    \includegraphics[width=\textwidth]
    {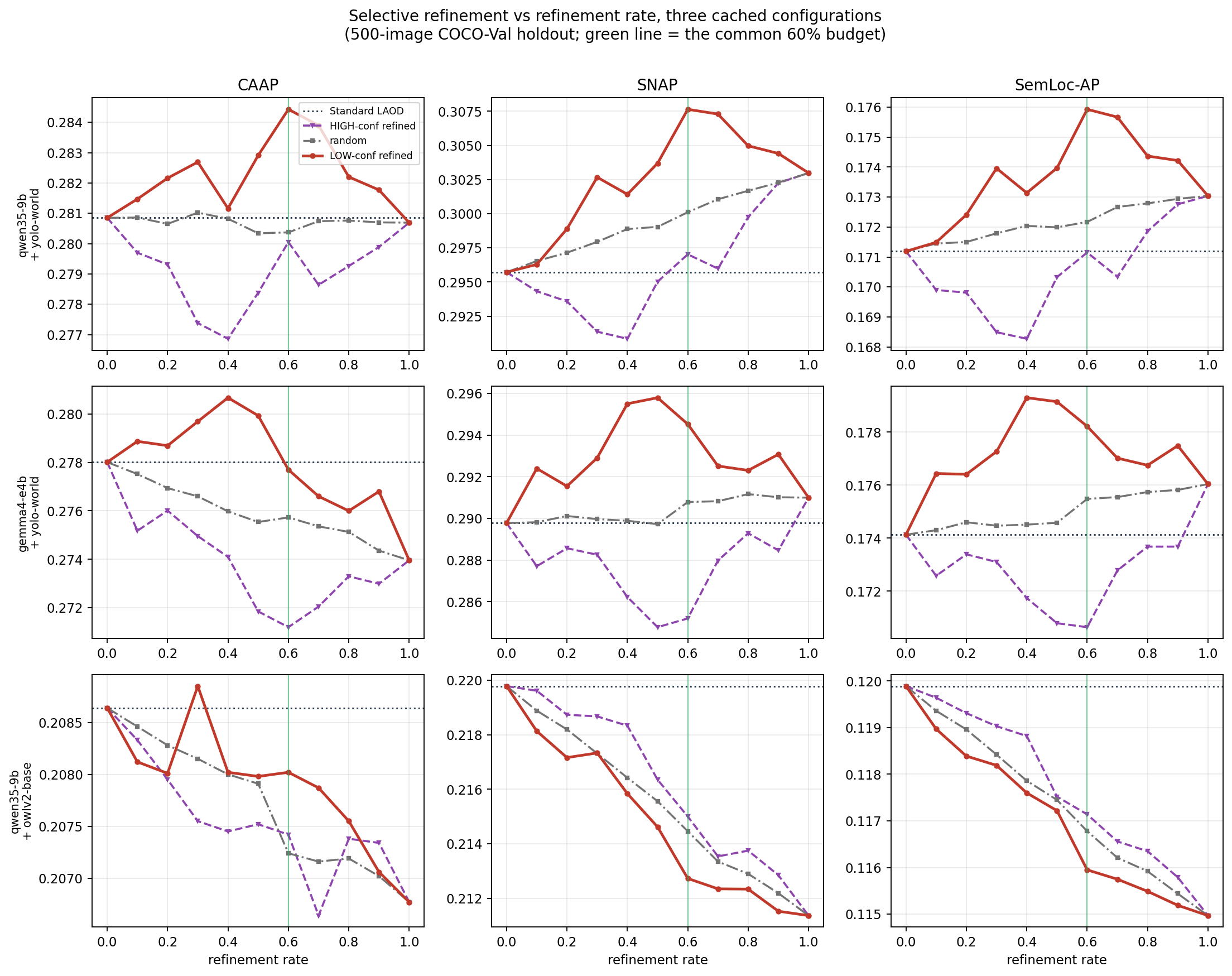}
    \caption{Effect of the refinement rate for the three evaluated
    MLLM--detector configurations. Images are ordered by first-pass maximum
    detector confidence, and the lowest-confidence fraction is refined.
    Qwen3.5-9B + YOLO-World exhibits a broad high-performing region around
    $55$--$75\%$, while Gemma4-E4B + YOLO-World reaches its strongest region
    earlier, around $40$--$50\%$. In contrast, Qwen3.5-9B + OWLv2 degrades
    as the refinement rate increases. The primary comparison uses a common
    $60\%$ budget rather than independently optimizing each configuration.}
    \label{fig:selective_refinement_curves}
\end{figure*}

The refinement-rate curves in
Figure~\ref{fig:selective_refinement_curves} reinforce this conclusion.
For Qwen3.5-9B with YOLO-World, the SemLoc-AP gain relative to one-pass
\LAOD{} increases from $+0.0003$ at a $10\%$ refinement rate to $+0.0047$
at $60\%$, before decreasing to $+0.0018$ under unconditional refinement.
Gemma4-E4B with YOLO-World follows a qualitatively similar but shifted pattern:
its SemLoc-AP gain reaches $+0.0050$ at $50\%$, decreases to $+0.0041$ at
$60\%$, and falls to $+0.0019$ when every image is refined. In contrast,
Qwen3.5-9B with OWLv2 degrades monotonically, with SemLoc-AP changes of
$-0.0009$, $-0.0027$, $-0.0039$, and $-0.0049$ at refinement rates of
$10\%$, $50\%$, $60\%$, and $100\%$, respectively. Thus, there is no
single refinement rate for which closing the loop is uniformly beneficial.

\paragraph{Does first-pass confidence provide a useful refinement gate?}
To determine whether low-confidence selection provides information beyond
simply refining an arbitrary subset of images, we compare it with both random
and high-confidence refinement at the same $60\%$ budget. Random refinement is
evaluated over 1,000 subsets. Table~\ref{tab:selective_refinement_controls}
reports the percentile attained by the low-confidence gate within the matched
random distribution.

\begin{table}[t]
\centering
\caption{Position of low-confidence selective refinement within the
1,000-run random-refinement distribution at the same $60\%$ refinement
budget. Higher percentiles indicate that selecting low-confidence images is
more effective than random selection.}
\label{tab:selective_refinement_controls}
\small
\setlength{\tabcolsep}{4.5pt}
\begin{tabular}{lccc}
\toprule
Configuration & CAAP & SNAP & SemLoc-AP \\
\midrule
Qwen3.5-9B + YOLO-World
& 99.2\% & 99.5\% & 99.1\% \\
Gemma4-E4B + YOLO-World
& 88.0\% & 95.2\% & 97.1\% \\
Qwen3.5-9B + OWLv2
& 72.6\% & 4.4\% & 4.1\% \\
\bottomrule
\end{tabular}
\end{table}

For both YOLO-World configurations, refining the lower-confidence subset
generally performs better than random refinement and substantially better than
refining the highest-confidence subset. Across these two configurations, the
ordering low-confidence $>$ random $>$ high-confidence refinement holds for
all six metric comparisons. The same behavior does not transfer to OWLv2.
Although CAAP remains above the random mean, the low-confidence gate falls at
only the $4.4$th and $4.1$st percentiles of the random distributions for SNAP
and SemLoc-AP, respectively. In these cases, selecting the lowest-confidence
images is actually worse than selecting an arbitrary $60\%$ of the dataset.
Across all three configurations and metrics, the
low-confidence $>$ random $>$ high-confidence ordering therefore holds in
only six of nine cases. The usefulness of detector confidence as a refinement
signal is itself detector dependent.

A secondary analysis considers whether the absolute YOLO-World threshold
selected from the Qwen3.5-9B configuration transfers across MLLMs. The
threshold $s_{\max}<0.9170$ refines $60.0\%$ of images for Qwen3.5-9B and
$52.4\%$ for Gemma4-E4B. Without retuning, the latter configuration reaches
CAAP $0.2795$, SNAP $0.2950$, and SemLoc-AP $0.1788$, corresponding to
changes of $+0.0015$, $+0.0052$, and $+0.0046$ relative to its one-pass
baseline. This indicates some transfer of the YOLO-World confidence scale
across MLLMs, but does not resolve the broader detector dependence observed
above.

\paragraph{Does the gate identify images that benefit from refinement?}
We additionally perform a retrospective help/hurt analysis using joint
semantic--spatial true positives at IoU $0.50$ and semantic threshold $0.60$.
Ground-truth information is used only for this post hoc diagnostic and never
by the refinement gate. For Qwen3.5-9B + YOLO-World, refinement helps 61
images, hurts 39, and leaves 400 unchanged under this diagnostic. The
corresponding counts are 45/69/386 for Gemma4-E4B + YOLO-World and
36/46/418 for Qwen3.5-9B + OWLv2.

First-pass maximum confidence does not consistently distinguish the images on
which refinement helps from those on which it hurts. For Qwen3.5-9B +
YOLO-World, the mean $s_{\max}$ values are $0.8341$ for helped examples and
$0.8272$ for hurt examples, with a helped-versus-hurt correlation of only
$0.023$. For Qwen3.5-9B + OWLv2, the corresponding means are $0.6198$ and
$0.6058$, with a correlation of $0.078$. Gemma4-E4B + YOLO-World is the
only configuration showing clearer separation, with mean values of $0.7315$
and $0.8210$ and a correlation of $-0.184$. Moreover, refinement hurts more
images than it helps for two of the three configurations. We therefore do not
interpret the confidence gate as identifying individual cases in which a
second MLLM--OVOD pass will be beneficial.

Instead, where selective refinement helps, its effect is more consistent with
\emph{damage avoidance}: additional interaction is withheld from stronger
first-pass outputs that are more likely to be perturbed unnecessarily. This
interpretation is supported by the high-confidence control, which performs
below random selection across the YOLO-World experiments. However, the failure
of the same strategy with OWLv2 shows that this behavior cannot be treated as
a general property of MLLM--OVOD interaction.

Finally, the computational cost of closing the loop is non-negligible.
Standard \LAOD{} requires one MLLM call and one OVOD pass per image. At the
common $60\%$ refinement budget, selective refinement increases both quantities
to $1.6$ per image, corresponding to $60\%$ more MLLM calls and $60\%$ more
detector passes. Unconditional refinement doubles both quantities. We report
these pass counts separately rather than combining them into a single runtime
factor because MLLM and detector inference have different computational costs.

Overall, these results do not support iterative MLLM--OVOD refinement as a
generally beneficial extension of \LAOD{}. Closing the loop can provide modest
semantic and joint semantic--spatial gains for some configurations, but the
effect is inconsistent across MLLMs and grounding models, can reduce
localization performance, and can become actively harmful for some detectors.
These gains must also be weighed against the additional MLLM and detector
passes required by refinement. We therefore retain the simple one-pass
formulation as the primary \LAOD{} design. Developing a more informative
criterion for determining \emph{when} the MLLM--OVOD loop should be invoked,
and \emph{how} detector feedback should modify the generated vocabulary,
remains an important direction for future work.

\section{Validation of the SNAP Semantic Encoder}
\label{app:snap_validation}

SNAP determines whether a predicted category name semantically matches a
ground-truth category by thresholding the cosine similarity between their text
embeddings. This requires an embedding space in which unrelated category names
are sufficiently separated; otherwise, low similarity thresholds accept
semantically unrelated labels and the metric ceases to measure naming quality.
We therefore evaluate the suitability of CLIP and Sentence-BERT as the text
encoder used by SNAP. The same semantic similarity criterion is also used by
SemLoc-AP.

All SNAP and SemLoc-AP results reported in this paper use Sentence-BERT
(\texttt{all-MiniLM-L6-v2}) and directly embed the bare category names. The
predicted label and ground-truth label are encoded independently, and their
cosine similarity is

\begin{equation}
s(\hat{l},l)
=
\frac{
\mathbf{e}(\hat{l})^\top \mathbf{e}(l)
}{
\|\mathbf{e}(\hat{l})\|_2
\|\mathbf{e}(l)\|_2
}.
\end{equation}

A semantic match is accepted when $s(\hat{l},l)\geq\tau$, where
$\tau\in\{0.50,0.55,\ldots,0.95\}$.

\subsection{CLIP Text Similarity}

Although CLIP provides text embeddings, its representation space is optimized
for image--text alignment rather than text--text semantic similarity. This
distinction is important for SNAP. Using CLIP ViT-B/32 with its standard
\texttt{``a photo of \{\}''} text template, we examine the cosine similarities
of all $3{,}160$ distinct pairs among the 80 COCO category names. Even
unrelated category names occupy a narrow region of the embedding space: the
minimum pairwise cosine similarity is $0.567$ and the median is $0.786$. For
example, the similarity between \emph{person} and \emph{car} is $0.883$.

Consequently, a semantic threshold of $\tau=0.50$ lies below the similarity of
every distinct COCO category pair under this representation. At this threshold,
CLIP accepts 100\% of the distinct category pairs. Thus, the semantic criterion
is effectively inactive: a prediction can satisfy the SNAP matching condition
regardless of whether its category name is semantically related to the ground
truth.

\begin{table}[t]
\centering
\caption{Behavior of CLIP and Sentence-BERT for SNAP on COCO-Val. The
``gain over shuffle'' measures the difference between SNAP using the original
predicted labels and SNAP after randomly permuting the predicted label strings
while keeping bounding boxes and confidence scores unchanged. Ranges summarize
the 28 MLLM--detector configurations.}
\label{tab:snap_encoder_validation}
\scriptsize
\setlength{\tabcolsep}{3.5pt}
\renewcommand{\arraystretch}{1.0}
\begin{tabular}{lcccc}
\toprule
Encoder
& LO gain
& MI gain
& HI gain
& Unrelated pairs at $\tau=.50$ \\
\midrule
CLIP ViT-B/32
& $0.0000$
& $0.0068$--$0.0861$
& $0.12$--$0.27$
& $100\%$ \\
Sentence-BERT
& $0.21$--$0.36$
& $0.17$--$0.28$
& $0.10$--$0.17$
& $2\%$ \\
\bottomrule
\end{tabular}
\end{table}

\subsection{Shuffled-Label Control}

To test whether SNAP actually depends on semantic information in the predicted
labels, we construct a shuffled-label control. The predicted bounding boxes,
detector confidence scores, and number of detections are kept exactly
unchanged, while the predicted category strings are randomly permuted across
the dataset. The shuffled predictions therefore preserve localization and
confidence information but remove the association between an object's
appearance and its predicted semantic name.

Table~\ref{tab:snap_encoder_validation} shows the resulting difference between
the original and shuffled-label scores. With CLIP, the LO SNAP score is
unchanged to four decimal places: the gain over shuffled labels is $0.0000$.
This behavior occurs for all 28 MLLM--detector combinations on COCO-Val. In
other words, under the LO thresholds, CLIP-based SNAP cannot distinguish the
actual predicted labels from randomly reassigned labels.

Sentence-BERT produces the opposite behavior. Only approximately $2\%$ of
distinct COCO category pairs exceed a similarity of $0.50$, and the original
predicted labels outperform their shuffled counterparts by $0.21$--$0.36$ in
the LO interval. The gain remains substantial at the MI and HI thresholds.
Sentence-BERT therefore yields informative semantic discrimination in all
28 of 28 evaluated COCO-Val configurations.

The difference is also visible in the absolute SNAP values. Replacing CLIP
with Sentence-BERT substantially lowers the LO scores because the semantic
criterion begins rejecting unrelated labels. This reduction should therefore
not be interpreted as degraded detection performance; rather, it removes the
component of the CLIP-based score that can be obtained without meaningful
category naming.

\subsection{Bare Category Names}

For the reported SNAP and SemLoc-AP metrics, we embed the category names
directly, e.g., \texttt{person}, \texttt{traffic light}, or
\texttt{teddy bear}, rather than wrapping every category in an image-oriented
prompt such as \texttt{``a photo of \{\}''}. The semantic criterion is intended
to measure agreement between the generated and annotated \emph{category names};
using bare labels therefore keeps the similarity computation directly tied to
those names rather than to shared prompt text.

This choice is also consistent with the role of Sentence-BERT, which is used
for text--text semantic comparison rather than image--text retrieval. The
threshold analysis and shuffled-label experiment above are consequently
performed with the same bare-label Sentence-BERT representation used for all
reported SNAP and SemLoc-AP results.

Overall, these experiments motivate the use of Sentence-BERT rather than CLIP
for the semantic matching component of our evaluation. CLIP's text embedding
geometry makes the lower semantic thresholds effectively non-discriminative,
whereas Sentence-BERT produces meaningful separation between related and
unrelated category names and retains a clear advantage over the shuffled-label
control throughout the evaluated threshold range.

\section{Detector Confidence Thresholds and Operating Points}
\label{app:detector_thresholds}

The confidence threshold of an open-vocabulary detector determines which
candidate detections are retained before evaluation. Because our primary
localization metric, CAAP, is based on average precision, we tune detector
thresholds using the same ranking-based protocol used for reporting results:
predictions are ranked by detector confidence and capped at a maximum of 100
detections per image before CAAP, SNAP, SemLoc-AP, or U-AP is computed. The
detector confidence is used as a ranking score and is not interpreted as a
globally calibrated probability. Consequently, numerical confidence values are
not directly comparable across detector families.

Detector thresholds are selected on the frozen 500-image hyperparameter
holdout. For COCO-Val and LVIS-minival, the tuning objective is CAAP with
$\mathrm{maxDets}=100$; for COCO-OOD, the objective is U-AP. Once selected,
a single threshold is fixed for each detector and used across datasets and
MLLMs. The selected thresholds are shown in
Table~\ref{tab:detector_thresholds}.

\begin{table}[t]
\centering
\caption{Detector confidence thresholds selected on the hyperparameter
holdout. A single threshold is fixed for each detector and used across all
evaluated datasets and MLLMs.}
\label{tab:detector_thresholds}
\scriptsize
\setlength{\tabcolsep}{4pt}
\renewcommand{\arraystretch}{0.95}
\begin{tabular}{lcc}
\toprule
Detector & Confidence parameter & Selected threshold \\
\midrule
YOLO-World & \texttt{conf} & 0.001 \\
G-DINO-T   & \texttt{score\_threshold} & 0.03 \\
G-DINO-B   & \texttt{score\_threshold} & 0.03 \\
OWLv2      & \texttt{score\_threshold} & 0.01 \\
\bottomrule
\end{tabular}
\end{table}

For both Grounding DINO variants, the text threshold is fixed at $0.3$ and is
not tuned. Across all 21 dataset--MLLM tuning combinations for each detector,
the selected confidence value is identical. In each case, the optimum occurs
at the lowest confidence threshold included in the corresponding sweep.

\subsection{Why Low Confidence Thresholds Favor AP}

This behavior follows directly from the ranking-based nature of average
precision. AP evaluates the precision--recall curve obtained by progressively
including detections in descending confidence order. Lowering the retention
threshold adds lower-ranked predictions without changing the ordering of
higher-confidence detections and may extend recall by recovering additional
true positives. Consequently, a low detector threshold is generally
appropriate when the final evaluation itself ranks detections by confidence.

The $\mathrm{maxDets}=100$ cap is important in this setting. Once predictions
are sorted by confidence, only the top 100 detections for an image can
contribute to the metric. Therefore, lowering the detector threshold below the
point needed to recover those top-ranked detections often has little or no
effect: additional predictions are generated but subsequently discarded by the
evaluation cap. In this regime, $\mathrm{maxDets}=100$ becomes the effective
final truncation mechanism.

This also explains why threshold selection should use the capped evaluation
objective. Optimizing uncapped CAAP can reward low-confidence detections that
are never considered by the final evaluation. For example, under the capped
objective OWLv2 selects a score threshold of $0.01$, whereas optimizing
uncapped CAAP selects $0.15$. The latter operating point is therefore
optimized under a prediction set that differs from the one used by the
reported evaluation protocol.

\subsection{AP and F1 Prefer Different Operating Points}

To illustrate the distinction between ranking-based AP and a single operating
point, Table~\ref{tab:yolo_threshold_sweep} re-scores the
Qwen3.5-9B + YOLO-World predictions over a range of confidence thresholds.
CAAP is highest at the lowest tested confidence threshold, whereas F1 peaks at
a substantially higher threshold.

\begin{table}[t]
\centering
\caption{Confidence-threshold sweep for Qwen3.5-9B + YOLO-World on COCO-Val.
F1, precision, and recall are computed at IoU $=0.5$. CAAP uses the full
IoU range from $0.50$ to $0.95$.}
\label{tab:yolo_threshold_sweep}
\scriptsize
\setlength{\tabcolsep}{3.8pt}
\renewcommand{\arraystretch}{0.93}
\begin{tabular}{ccccc}
\toprule
Conf. & CAAP & F1 & Precision & Recall \\
\midrule
0.001 & \textbf{0.3030} & 0.1770 & 0.100 & \textbf{0.803} \\
0.005 & 0.3003 & 0.2545 & 0.152 & 0.767 \\
0.010 & 0.2974 & 0.3043 & 0.192 & 0.740 \\
0.020 & 0.2925 & 0.3611 & 0.243 & 0.705 \\
0.030 & 0.2883 & 0.3954 & 0.279 & 0.678 \\
0.050 & 0.2816 & 0.4340 & 0.329 & 0.639 \\
0.075 & 0.2735 & 0.4593 & 0.372 & 0.601 \\
0.100 & 0.2657 & 0.4721 & 0.403 & 0.570 \\
0.150 & 0.2515 & \textbf{0.4765} & 0.445 & 0.513 \\
0.200 & 0.2384 & 0.4729 & 0.476 & 0.470 \\
0.250 & 0.2266 & 0.4673 & 0.504 & 0.436 \\
0.300 & 0.2179 & 0.4601 & 0.529 & 0.407 \\
0.400 & 0.1996 & 0.4419 & 0.575 & 0.359 \\
0.500 & 0.1836 & 0.4158 & \textbf{0.612} & 0.315 \\
\bottomrule
\end{tabular}
\end{table}

The sweep shows the expected precision--recall trade-off. Raising the
confidence threshold removes lower-ranked detections, increasing precision but
steadily reducing recall. F1 therefore reaches its maximum at an intermediate
threshold, $0.15$, where precision and recall are relatively balanced. CAAP
behaves differently: it is highest at $0.001$ and decreases monotonically as
the confidence threshold increases.

This distinction is important because F1 and AP answer different evaluation
questions. F1 evaluates performance at a single selected operating point,
whereas AP evaluates the quality of the confidence ranking across operating
points. Using an F1-optimal detector threshold for an AP-based benchmark would
prematurely remove lower-ranked detections that can contribute additional
recall. We consequently select detector thresholds using the same AP-style
objectives employed by the main evaluation and keep the resulting threshold
fixed for each detector across datasets and MLLMs.

\section{Open-Ended Detection Baseline Details}
\label{app:direct_detection}

The main paper compares \LAOD{} against two complementary open-ended detection
baselines. GenerateU~\cite{lin2024generateu} is a detector developed and
trained specifically for open-ended object detection, whereas the direct MLLM
baseline uses the same vision-capable MLLMs evaluated in \LAOD{} to jointly
generate object names and bounding-box coordinates. This appendix provides the
implementation and evaluation details needed to make both comparisons
consistent with the ontology-free \LAOD{} evaluation protocol.

\subsection{GenerateU Configuration}

We evaluate the strongest publicly released GenerateU checkpoint,
\texttt{vg\_grit5m\_swinL.pth}, which uses a Swin-L backbone and was trained
using Visual Genome and GRIT-5M. The official GenerateU model configuration is
used without modification.

GenerateU jointly localizes image regions and generates free-form semantic
names rather than requiring a predefined category vocabulary at inference time.
It therefore provides a direct task-level comparison to \LAOD{}, although the
two methods address open-ended detection differently. GenerateU is trained as a
unified open-ended detector, whereas \LAOD{} composes separately pretrained
MLLMs and open-vocabulary detectors through a textual vocabulary interface.

GenerateU is evaluated on the same frozen test splits used throughout the main
paper: 4,500 COCO-Val images containing 32,814 annotated objects and 4,327
LVIS-minival images containing 45,063 annotated objects.

\subsection{Ontology-Free GenerateU Evaluation}

The released GenerateU inference procedure incorporates the benchmark ontology
into its final prediction score and category assignment. Specifically, after
generating a free-form name for a localized region, GenerateU computes
similarity between the generated text and LVIS category embeddings. Its
reported confidence combines the region-level objectness and predicted
localization quality with this category-similarity term, while the final label
is replaced by the corresponding LVIS category index.

Using this final score would introduce benchmark-ontology information into an
otherwise open-ended evaluation. We therefore stop before the category-dependent
mapping step. Each prediction retains its generated free-form object name and
bounding box, and detections are ranked using the native
objectness--IoU-quality product computed immediately before the LVIS similarity
term is introduced. No CLIP- or BERT-based remapping to COCO or LVIS category
names is performed.

This preserves the same ontology-free setting used for \LAOD{}: neither method
is provided with the benchmark category set when generating its semantic
predictions.

\subsection{GenerateU Label and Prediction Handling}

GenerateU decodes multiple textual beam hypotheses for each predicted region
and returns them as a comma-separated sequence, for example
\emph{``bear, bears, bears fur''}. CAAP, SNAP, and SemLoc-AP assume a single
semantic label for each detection. We therefore use the first generated beam
hypothesis as the predicted object name, yielding \emph{``bear''} in this
example. The resulting label remains a free-form GenerateU prediction and is
not projected onto the benchmark ontology.

The released configuration uses
\texttt{TEST.NUM\_TEST\_QUERIES=300} and applies no score threshold, resulting
in exactly 300 candidate detections for every image. In the default GenerateU
inference path, the NMS branch is executed only when the number of instances
exceeds 300; consequently, no NMS is applied under this configuration. We
introduce no additional prediction filtering.

As for all other methods in our evaluation, $\mathrm{maxDets}=100$ is applied
exactly once per image before computing CAAP, SNAP, and SemLoc-AP. Predictions
are retained before this step so that the common evaluation code, rather than
the baseline-specific inference pipeline, performs the final truncation.

\subsection{Direct MLLM Prompt and Decoding}

The same direct-detection prompt is used for all evaluated models:

\begin{quote}
\small
\texttt{Detect all objects in this image and provide their bounding box
coordinates. Output the results strictly in a JSON list format using
'bbox\_2d' and 'label'.}
\end{quote}

Generation is performed greedily with
\texttt{max\_new\_tokens=1024}. Each valid output is parsed into a set of
predicted bounding boxes and corresponding category names. No
open-vocabulary detector is used in this baseline.

\subsection{Ranking Direct Predictions}

For the direct MLLM baseline, CAAP, SNAP, SemLoc-AP, and U-AP require
detections to be ranked by confidence. Direct MLLM outputs, however, do not
provide detector confidence scores analogous to those produced by YOLO-World,
Grounding DINO, or OWLv2. We therefore use token log-probabilities available
during generation as a ranking score.

Specifically, each predicted box is assigned the mean token log-probability of
the text representing its coordinates. Let $\mathcal{B}_i$ denote the set of
generated tokens corresponding to the coordinates of prediction $i$. Its
ranking score is

\begin{equation}
s_i =
\frac{1}{|\mathcal{B}_i|}
\sum_{t \in \mathcal{B}_i}
\log p(y_t \mid y_{<t}, X, p),
\end{equation}

where $X$ is the input image and $p$ is the direct-detection prompt.
Predictions are ranked in descending order of $s_i$. We use this
\texttt{logprob\_bbox} score for all reported metrics so that the same
prediction ordering is used for CAAP, SNAP, SemLoc-AP, and U-AP. This score is
used only for ranking and is not interpreted as a calibrated detection
probability or as being numerically comparable to the confidence scores
produced by the open-vocabulary detectors.

Token log-probabilities must be recorded during generation and cannot in
general be reconstructed from the final textual response. Accordingly, the
bounding-box score is stored together with each parsed prediction.

\subsection{Coordinate Conversion}

The evaluated models use different native coordinate conventions, requiring
model-specific conversion before evaluation. Qwen2.5-VL produces coordinates
in the resized image frame processed by the model. We recover the dimensions
of this frame from the processor output, including \texttt{image\_grid\_thw},
and rescale the predicted coordinates back to the original image dimensions.

InternVL3 instead represents box coordinates on a normalized $0$--$1000$
scale. For an image with width $W$ and height $H$, an InternVL3 prediction
$(x_1,y_1,x_2,y_2)$ is converted as

\begin{equation}
\begin{aligned}
x'_1 &= \frac{x_1}{1000}W, \qquad
x'_2 = \frac{x_2}{1000}W,\\
y'_1 &= \frac{y_1}{1000}H, \qquad
y'_2 = \frac{y_2}{1000}H.
\end{aligned}
\end{equation}

The coordinate conventions were verified before running the full evaluation.
For InternVL3, for example, the correct normalization produced a mean
best-IoU of $0.480$ against ground truth, compared with $0.291$ under the
incorrect coordinate interpretation. We therefore treat the coordinate format
as model-specific rather than attempting to apply a common default conversion.

\subsection{Parsing Reliability}

Direct detection additionally depends on the model following the requested
structured-output format. Table~\ref{tab:direct_parse_failures} reports the
number of images for which no valid detection list could be recovered from the
generated response.

\begin{table}[t]
\centering
\caption{Parsing statistics for the direct-detection baselines. A parsing
failure indicates that the generated response could not be converted into the
required bounding-box and label representation.}
\label{tab:direct_parse_failures}
\scriptsize
\setlength{\tabcolsep}{5pt}
\renewcommand{\arraystretch}{0.95}
\begin{tabular}{lcc}
\toprule
Model & COCO-Val & COCO-OOD \\
\midrule
Qwen2.5-VL-7B & 70  & 0 \\
InternVL3-8B   & 328 & 1 \\
\bottomrule
\end{tabular}
\end{table}

The difference is substantial on COCO-Val: Qwen2.5-VL fails to produce a
parseable output for 70 images, whereas InternVL3 fails on 328. On COCO-OOD,
both models follow the requested format substantially more reliably, with zero
and one parsing failures, respectively. These failures are part of the direct
baseline itself: unlike \LAOD{}, where localization is returned in the native
structured format of a detector, direct MLLM detection requires free-form
generation to satisfy a precise coordinate schema.

\subsection{Qwen3.5 Direct Detection}

We also evaluated Qwen3.5-9B as a direct detector. In this setting, the model
did not reliably follow the requested JSON-only detection format and produced
zero parseable detections across 1,798 attempted images. Disabling its
reasoning mode through the available generation interface did not resolve this
behavior. We therefore exclude Qwen3.5-9B from the quantitative
direct-detection comparison rather than assigning zero detection performance
to malformed outputs.

This issue is specific to using the model as a coordinate-generating detector.
Qwen3.5-9B remains fully usable within \LAOD{}, where its task is to generate
an image-specific list of object concepts while localization is handled by the
specialized open-vocabulary detector.

\section{Fixed-Ontology Reference: Removing Vocabulary Discovery}
\label{app:fixed_ontology}

The primary \LAOD{} experiments require the MLLM to construct an
image-specific detection vocabulary before spatial grounding. To isolate the
effect of this vocabulary-discovery stage, we additionally evaluate the
open-vocabulary detectors when they are provided with the complete benchmark
ontology directly. Specifically, every COCO-Val image is queried using all
80 COCO category names, while every LVIS-minival image is queried using all
1,203 LVIS category names. No MLLM is used in this experiment.

This setting should not be interpreted as an upper bound, oracle, or directly
comparable autonomous baseline. Supplying the benchmark ontology provides
privileged knowledge of the category space and removes most of the
vocabulary-discovery requirement. The detector must nevertheless determine
which categories are present, enumerate instances, localize them, rank the
resulting predictions, and suppress false positives and duplicates. We
therefore treat the experiment as a \emph{fixed-ontology reference} for
understanding how vocabulary discovery interacts with spatial grounding.

\subsection{Protocol}

We use the same frozen evaluation splits as in the main experiments:
4,500 COCO-Val images containing 32,814 annotated objects and 4,327
LVIS-minival images containing 45,063 annotated objects. In both cases, the
500-image hyperparameter-selection holdout is excluded.

For each dataset, category names are taken directly from the annotation
ontology and ordered deterministically by category identifier. The same
normalization used by the evaluation pipeline is applied to both predicted and
ground-truth category names. For LVIS, this converts underscores to spaces in
412 of the 1,203 category names; COCO names are unchanged. We do not introduce
synonyms, aliases, definitions, descriptions, or additional prompt
engineering.

All detector settings are kept identical to the main experiments and are not
retuned for the fixed-ontology setting. YOLO-World uses a confidence threshold
of $0.001$, OWLv2 uses a score threshold of $0.01$, and both Grounding DINO
variants use a score threshold of $0.03$ together with a text threshold of
$0.3$.

The LVIS ontology introduces an additional implementation challenge for
Grounding DINO. The complete 1,203-category ontology requires 3,802 text
tokens, exceeding its 256-token text-encoder capacity. We therefore divide the
ontology into 16 deterministic chunks, with every category appearing exactly
once. Each chunk is processed independently on the same image, after which the
resulting detections are concatenated while preserving their predicted labels
and confidence scores. Predictions are then globally ranked, and the common
$\mathrm{maxDets}=100$ limit is applied once after all chunks have been merged.
In contrast, YOLO-World consumes all 1,203 categories in a single
\texttt{set\_classes} call, and all LVIS category queries fit within OWLv2's
per-query text limit, so neither detector requires chunking.

All results are evaluated using the same authoritative metrics as the main
paper: CAAP with the CAAP-V2 protocol, SNAP using
Sentence-BERT (\texttt{all-MiniLM-L6-v2}) over bare category names, and
SemLoc-AP over the full $10\times10$ Cartesian product of spatial and semantic
thresholds. The same final prediction set, capped at 100 detections per image,
is used for all three metrics.

\subsection{COCO-Val: Fixed COCO-80 Ontology}

Table~\ref{tab:fixed_ontology_coco} reports the fixed-ontology results on
COCO-Val. All four detectors process the complete COCO-80 ontology in a single
text pass.

\begin{table*}[t]
\centering
\caption{Fixed-ontology reference on COCO-Val. Every image is queried using
the complete COCO-80 category set. ``Pre-cap'' reports the mean number of
candidate detections per image before the common $\mathrm{maxDets}=100$
truncation, while ``Post-cap'' reports the mean number retained afterward.}
\label{tab:fixed_ontology_coco}
\small
\setlength{\tabcolsep}{6pt}
\begin{tabular}{lcccccc}
\toprule
Detector & Vocabulary & CAAP & SNAP & SemLoc-AP &
Det./Img Pre-cap & Det./Img Post-cap \\
\midrule
\textbf{Grounding DINO-B}
& COCO-80 & \textbf{0.5559} & \textbf{0.8803} & \textbf{0.5408}
& 190.1 & 75.5 \\

Grounding DINO-T
& COCO-80 & 0.4935 & 0.7815 & 0.4688
& 220.8 & 81.1 \\

YOLO-World
& COCO-80 & 0.4916 & 0.7958 & 0.4686
& 103.8 & 63.4 \\

OWLv2
& COCO-80 & 0.3289 & 0.5925 & 0.3072
& 227.2 & 83.3 \\
\bottomrule
\end{tabular}
\end{table*}

Providing the complete COCO ontology substantially improves all three metrics.
Grounding DINO-B achieves the strongest result, reaching a CAAP of $0.5559$,
SNAP of $0.8803$, and SemLoc-AP of $0.5408$. Grounding DINO-T and
YOLO-World obtain similar localization and joint performance, with CAAP values
of $0.4935$ and $0.4916$ and SemLoc-AP values of $0.4688$ and $0.4686$,
respectively. YOLO-World obtains the higher SNAP of the two at $0.7958$,
compared with $0.7815$ for Grounding DINO-T. OWLv2 also benefits from the
fixed ontology but remains the weakest detector on COCO, reaching $0.3289$
CAAP, $0.5925$ SNAP, and $0.3072$ SemLoc-AP.

The particularly large SNAP values are expected. In this setting, the detector
is queried directly using the benchmark's canonical category names, removing
synonym variation, granularity differences, omitted concepts, and hallucinated
concepts from the vocabulary-generation stage. Nevertheless, the improvements
are not restricted to naming: CAAP and SemLoc-AP also increase substantially,
showing that the supplied vocabulary affects which objects are successfully
grounded and how their detections are ranked.

\subsection{LVIS-minival: Fixed LVIS-1203 Ontology}

Table~\ref{tab:fixed_ontology_lvis} reports the corresponding experiment on
LVIS-minival. Unlike COCO, the complete LVIS ontology contains 1,203
fine-grained categories and produces substantially different behavior across
detectors.

\begin{table*}[t]
\centering
\caption{Fixed-ontology reference on LVIS-minival using the complete
1,203-category LVIS ontology. Grounding DINO requires 16 independent text
passes because the ontology exceeds its text-encoder capacity. The pre-cap
counts for the Grounding DINO variants therefore aggregate detections across
all 16 passes and are not directly comparable with the single-pass counts of
YOLO-World and OWLv2.}
\label{tab:fixed_ontology_lvis}
\small
\setlength{\tabcolsep}{5pt}
\begin{tabular}{lccccccc}
\toprule
Detector & Vocabulary & CAAP & SNAP & SemLoc-AP &
Det./Img Pre-cap & Det./Img Post-cap & Chunks \\
\midrule
\textbf{OWLv2}
& LVIS-1203 & \textbf{0.2461} & \textbf{0.3434} & \textbf{0.1801}
& 603.9 & 98.7 & 1 \\

YOLO-World
& LVIS-1203 & 0.1639 & 0.1731 & 0.0976
& 238.0 & 97.0 & 1 \\

Grounding DINO-B
& LVIS-1203 & 0.1031 & 0.1346 & 0.0769
& 912.7 & 99.9 & 16 \\

Grounding DINO-T
& LVIS-1203 & 0.0689 & 0.0696 & 0.0418
& 11,745.9 & 100.0 & 16 \\
\bottomrule
\end{tabular}
\end{table*}

The fixed ontology provides a much less uniform advantage on LVIS. OWLv2
benefits most clearly, reaching $0.2461$ CAAP, $0.3434$ SNAP, and $0.1801$
SemLoc-AP. YOLO-World reaches $0.1639$, $0.1731$, and $0.0976$,
respectively. In contrast, both Grounding DINO variants perform substantially
worse than the strongest single-pass detectors. As discussed below, however,
the Grounding DINO LVIS results are confounded by the need to process the
ontology in multiple independent text passes and should not be interpreted as
clean fixed-ontology reference values.

\subsection{Comparison with Autonomous Vocabulary Generation}

Table~\ref{tab:fixed_vs_autonomous} compares each fixed-ontology result against
the corresponding detector average under autonomous \LAOD{} vocabulary
generation. The autonomous values average across the MLLMs evaluated with each
detector in the main experiments.

\begin{table*}[t]
\centering
\caption{Comparison between autonomous \LAOD{} detector averages and the
fixed-ontology reference. Positive values indicate improvement when the
complete benchmark ontology is supplied directly. The Grounding DINO LVIS
comparisons are affected by the multi-pass limitation discussed in
Section~\ref{app:fixed_ontology_gdino}.}
\label{tab:fixed_vs_autonomous}
\scriptsize
\setlength{\tabcolsep}{4pt}
\resizebox{\textwidth}{!}{
\begin{tabular}{llcccccc}
\toprule
Dataset & Detector &
CAAP Auto $\rightarrow$ Fixed & $\Delta$CAAP &
SNAP Auto $\rightarrow$ Fixed & $\Delta$SNAP &
SemLoc Auto $\rightarrow$ Fixed & $\Delta$SemLoc \\
\midrule

COCO
& YOLO-World
& $0.2882 \rightarrow 0.4916$ & $+0.2034$
& $0.2899 \rightarrow 0.7958$ & $+0.5059$
& $0.1779 \rightarrow 0.4686$ & $+0.2907$ \\

& G-DINO-B
& $0.2634 \rightarrow 0.5559$ & $+0.2925$
& $0.2097 \rightarrow 0.8803$ & $+0.6706$
& $0.1270 \rightarrow 0.5408$ & $+0.4138$ \\

& G-DINO-T
& $0.2522 \rightarrow 0.4935$ & $+0.2413$
& $0.2179 \rightarrow 0.7815$ & $+0.5636$
& $0.1342 \rightarrow 0.4688$ & $+0.3346$ \\

& OWLv2
& $0.2129 \rightarrow 0.3289$ & $+0.1160$
& $0.2159 \rightarrow 0.5925$ & $+0.3766$
& $0.1249 \rightarrow 0.3072$ & $+0.1823$ \\

\midrule

LVIS
& YOLO-World
& $0.1645 \rightarrow 0.1639$ & $-0.0006$
& $0.1687 \rightarrow 0.1731$ & $+0.0044$
& $0.0913 \rightarrow 0.0976$ & $+0.0063$ \\

& OWLv2
& $0.1963 \rightarrow 0.2461$ & $+0.0498$
& $0.2159 \rightarrow 0.3434$ & $+0.1275$
& $0.1172 \rightarrow 0.1801$ & $+0.0629$ \\

& G-DINO-B
& $0.1215 \rightarrow 0.1031$ & $-0.0184$
& $0.1272 \rightarrow 0.1346$ & $+0.0074$
& $0.0689 \rightarrow 0.0769$ & $+0.0080$ \\

& G-DINO-T
& $0.1451 \rightarrow 0.0689$ & $-0.0762$
& $0.1408 \rightarrow 0.0696$ & $-0.0712$
& $0.0797 \rightarrow 0.0418$ & $-0.0379$ \\

\bottomrule
\end{tabular}
}
\end{table*}

The difference between the two datasets is substantial. On COCO, removing the
vocabulary-discovery requirement improves CAAP by $0.1160$--$0.2925$, SNAP by
$0.3766$--$0.6706$, and SemLoc-AP by $0.1823$--$0.4138$ across the four
detectors. Grounding DINO-B shows the largest gains, increasing from $0.2634$
to $0.5559$ CAAP, from $0.2097$ to $0.8803$ SNAP, and from $0.1270$ to
$0.5408$ SemLoc-AP.

The gains are particularly large for SNAP, which is expected because the
fixed-ontology setting supplies the detector with the exact canonical benchmark
names. The semantic naming problem is therefore greatly simplified by
construction. The simultaneous CAAP gains show, however, that vocabulary
selection also affects localization: providing a category to the grounding
model changes whether corresponding object instances are proposed and how they
are ranked. SemLoc-AP captures both effects and consequently improves
substantially as well.

The LVIS behavior is markedly different. OWLv2 remains the strongest detector
and benefits substantially from the fixed ontology, with gains of $+0.0498$
CAAP, $+0.1275$ SNAP, and $+0.0629$ SemLoc-AP. YOLO-World, however, obtains
almost identical localization performance under autonomous and fixed
vocabularies: CAAP changes by only $-0.0006$, while SNAP and SemLoc-AP improve
by $+0.0044$ and $+0.0063$. Thus, for YOLO-World, supplying all 1,203
benchmark categories provides essentially no localization advantage over the
compact image-specific vocabularies generated by the MLLM.

The detector ranking also changes when vocabulary discovery is removed. On
COCO, YOLO-World has the highest average CAAP under autonomous vocabulary
generation ($0.2882$), whereas Grounding DINO-B becomes the strongest detector
under the fixed ontology ($0.5559$). Grounding DINO-T and YOLO-World become
nearly tied under the fixed ontology. On LVIS, OWLv2 remains the strongest
detector in both settings, but its advantage increases substantially when the
complete ontology is supplied. These changes show that detector performance
under autonomous, compact vocabularies does not directly predict behavior under
a large known category set.

\subsection{Grounding DINO Limitation on the Full LVIS Ontology}
\label{app:fixed_ontology_gdino}

The Grounding DINO LVIS results require additional qualification because the
complete ontology cannot be processed in a single text pass. The 1,203 LVIS
categories require 3,802 text tokens, while the Grounding DINO text encoder
supports only 256. The ontology is therefore divided into 16 independent
chunks, each of which requires a separate forward pass.

This produces two related problems. First, the number of retained candidates
increases dramatically. Grounding DINO-B produces an average of approximately
$912.7$ detections per image before the common maxDets cap, whereas
Grounding DINO-T produces approximately $11{,}745.9$. For Grounding DINO-T,
a per-chunk trace shows between 732 and 859 detections in nearly every one of
the 16 chunks. The median confidence within each chunk lies between $0.045$
and $0.066$, above the fixed $0.03$ detector threshold, causing a near-uniform
flood of candidates. Grounding DINO-B remains substantially more selective,
producing approximately 1--18 candidates per chunk under the same threshold.

Second, predictions produced by different chunks arise from independent
forward passes. Their confidence scores must nevertheless be concatenated and
placed into a single global ranking before applying maxDets. The per-chunk
maximum scores vary substantially, meaning that this procedure implicitly
assumes that confidence values from independent vocabulary chunks share a
common ranking scale. With Grounding DINO-T in particular, the resulting
top-100 predictions can therefore be dominated by the chunks that produce
larger numerical scores rather than by the categories most relevant to the
image.

\begin{table}[t]
\centering
\caption{Observed Grounding DINO behavior when the complete LVIS ontology is
processed in 16 independent vocabulary chunks.}
\label{tab:gdino_lvis_chunking}

\small
\setlength{\tabcolsep}{3pt}

\begin{tabular}{lccc}
\toprule
Detector &
\makecell{Boxes/\\Chunk} &
\makecell{Per-Chunk\\Max Score} &
\makecell{Per-Chunk\\Median Score} \\
\midrule

G-DINO-T
& 732--859
& 0.247--0.608
& 0.045--0.066 \\

G-DINO-B
& 1--18
& 0.087--0.793
& 0.034--0.106 \\

\bottomrule
\end{tabular}
\end{table}

We deliberately retain the detector thresholds used throughout the main
experiments rather than retuning them specifically for this fixed-ontology
reference. Retuning could change the resulting candidate distribution, but
would introduce a different detector configuration and make the comparison
dependent on ontology-specific hyperparameter optimization. We therefore
report the Grounding DINO LVIS results for completeness while treating them as
evidence of a detector-specific scalability limitation rather than as clean
fixed-ontology reference values.

This limitation is itself informative. A fixed vocabulary is straightforward
when the ontology is compact, as in COCO-80, but supplying a very large
category set can exceed the native text capacity of the grounding model and
require multiple independently scored forward passes. In such cases, the cost
and ranking behavior of the detector become part of the problem rather than
vocabulary discovery alone.

\subsection{Discussion}

The fixed-ontology experiments reveal a clear interaction between ontology
size, vocabulary discovery, and grounding. On COCO, the vocabulary-discovery
gap is large. Providing the 80 canonical categories improves all detectors,
with particularly large gains in semantic naming. The fixed ontology removes
synonyms, granularity mismatches, omissions, and hallucinated concepts from the
vocabulary-generation stage, allowing the grounding models to operate directly
on the benchmark's category names.

The same conclusion does not transfer directly to LVIS. For OWLv2, the
complete ontology remains beneficial, particularly for semantic naming. For
YOLO-World, however, supplying all 1,203 categories provides almost no
localization improvement over autonomous image-specific vocabulary generation.
The Grounding DINO variants additionally encounter substantial practical
difficulties because the ontology exceeds their text capacity and must be
processed in multiple independently scored chunks.

The results therefore suggest that the usefulness of a fixed ontology depends
strongly on both the size of the category space and the grounding model. A
compact known ontology such as COCO-80 can substantially simplify the task.
For a large long-tail ontology such as LVIS-1203, however, considering every
possible category for every image can itself create grounding, ranking, and
scalability challenges. Even the strongest fixed-ontology LVIS configuration,
OWLv2, reaches a SemLoc-AP of only $0.1801$, compared with $0.5408$ for
Grounding DINO-B on COCO.

These results also clarify the role of image-specific vocabulary generation in
\LAOD{}. Autonomous vocabulary discovery introduces semantic uncertainty, as
shown by the large COCO gap between autonomous and fixed vocabularies. At the
same time, it provides the grounding model with a compact set of concepts
selected from the visual content rather than requiring the detector to consider
the entire benchmark ontology. This becomes particularly relevant as the
category space grows. The fixed-ontology results should therefore be viewed as
an ontology-aware reference that isolates vocabulary discovery, rather than as
a universal ceiling on autonomous detection performance.

\section{Additional Qualitative Examples Across Benchmarks}
\label{app:additional_qualitative}

\begin{figure*}[t]
    \centering

    \begin{subfigure}[t]{0.32\linewidth}
        \centering
        \qualimg{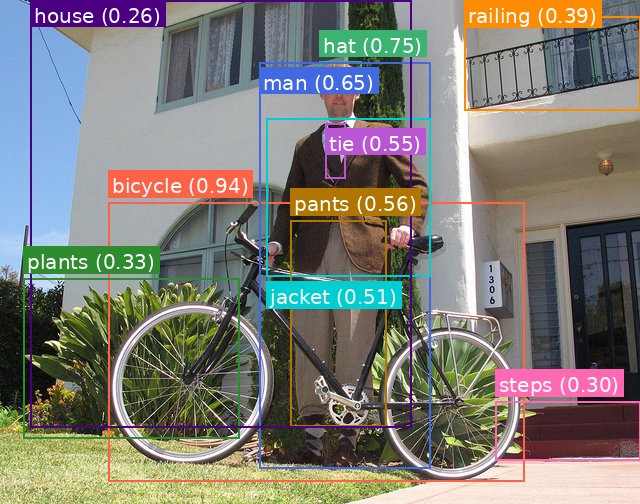}\\[-1mm]
        {\scriptsize \LAOD{} predictions}\\[1mm]
        \qualimg{}\\[-1mm]
        {\scriptsize Ground truth}
        \caption{COCO-Val}
    \end{subfigure}
    \hfill
    \begin{subfigure}[t]{0.32\linewidth}
        \centering
        \qualimg{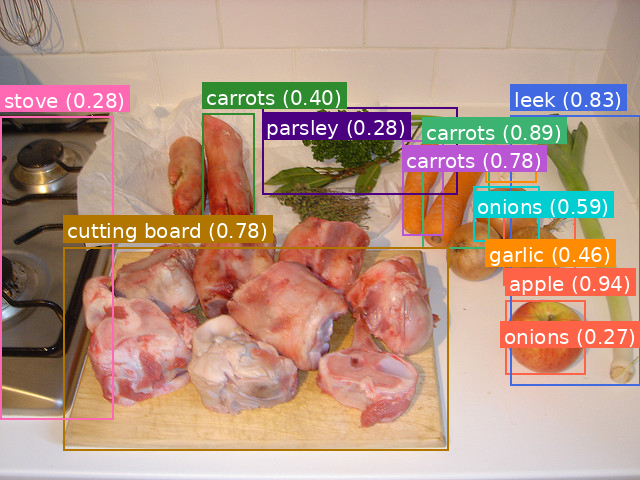}\\[-1mm]
        {\scriptsize \LAOD{} predictions}\\[1mm]
        \qualimg{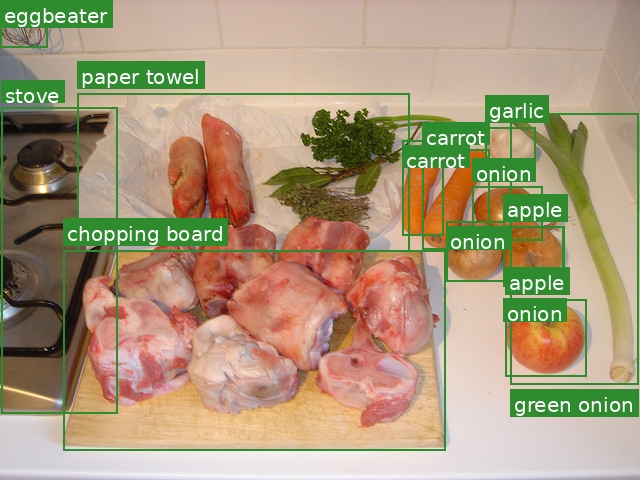}\\[-1mm]
        {\scriptsize Ground truth}
        \caption{LVIS-minival}
    \end{subfigure}
    \hfill
    \begin{subfigure}[t]{0.32\linewidth}
        \centering
        \qualimg{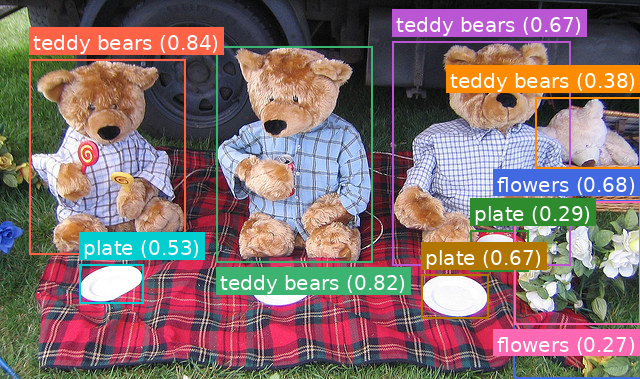}\\[-1mm]
        {\scriptsize \LAOD{} predictions}\\[1mm]
        \qualimg{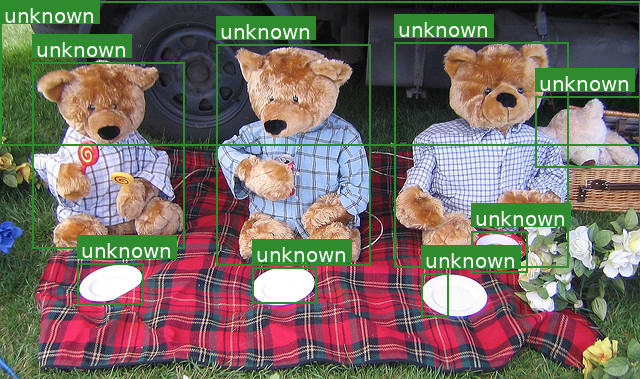}\\[-1mm]
        {\scriptsize Ground truth}
        \caption{COCO-OOD}
    \end{subfigure}

    \caption{Qualitative examples across the three evaluation settings.
    On COCO-Val, \LAOD{} localizes the annotated objects while also identifying
    additional plausible concepts outside the benchmark annotations. On
    LVIS-minival, it recognizes fine-grained objects and semantically related
    names, such as \emph{cutting board} for the annotated \emph{chopping board}.
    On COCO-OOD, \LAOD{} assigns meaningful names such as \emph{teddy bears},
    \emph{plates}, and \emph{flowers}, whereas the benchmark represents these
    objects using the generic \emph{unknown} label.}
    \label{fig:qualitative}
\end{figure*}

Figure~\ref{fig:qualitative} provides additional qualitative examples from
COCO-Val, LVIS-minival, and COCO-OOD. The examples illustrate behaviors that
are difficult to capture fully with fixed-vocabulary evaluation: plausible
detections outside the annotated ontology, semantically related names that
differ from the canonical annotation terminology, and meaningful semantic
labels for objects represented only by the generic \emph{unknown} category in
COCO-OOD.

\section{Discussion of Potential Applications}

The ability to construct an image-specific detection vocabulary without
requiring users to enumerate the target categories may be useful in downstream
systems operating in environments where the relevant objects cannot be fully
specified in advance. One example is video anomaly detection, where object
detectors are frequently used to represent scene content and identify
potentially anomalous entities or interactions~\cite{mumcu2025complexvad, mumcu2025leveraging}. A fixed detector vocabulary can limit such systems when an
anomaly involves an object category that was not anticipated during detector
training. \LAOD{} could instead provide scene-dependent object detections that
can be passed to subsequent anomaly-reasoning modules.

Similar requirements arise in autonomous robotics, surveillance, and
industrial inspection~\cite{liu2024okrobot,rahman2024opennav,
pham2024fisheye,wang2026industrial}. Robots and autonomous systems may encounter previously
unanticipated objects or obstacles, security systems may need to reason about
scene-specific items rather than a fixed list of categories, and industrial
inspection systems may operate across facilities containing different
components and foreign objects. In these settings, the modular design of
\LAOD{} provides a mechanism for dynamically determining what concepts should
be detected while retaining the localization capabilities of specialized
open-vocabulary detectors. These are potential downstream uses rather than
tasks evaluated in this work, and their effectiveness would require
application-specific validation.

\section{Additional Implementation Details}
\label{app:implementation_details}

\subsection{Dataset Splits}

We use a frozen 500-image hyperparameter-selection holdout drawn from COCO
val2017 with seed 20261001. Image identifiers are sorted before sampling, and
the resulting holdout is fixed once generated. The reported COCO-Val test split
is defined as the complement of this holdout rather than as an independently
sampled subset.

The same holdout exclusion is applied consistently across all three benchmarks.
Because LVIS-minival and COCO-OOD contain images drawn from COCO val2017, some
of their images also belong to the holdout and are therefore removed from
evaluation. After this exclusion, the reported test sets contain 4,500
COCO-Val images with 32,814 annotated objects, 4,327 LVIS-minival images with
45,063 annotated objects, and 438 COCO-OOD images with 1,386 annotated
objects.

The prompt-ablation experiments use separate frozen 500-image subsets sampled
from the corresponding test sets after removal of the hyperparameter holdout.
The COCO subset uses seed 20261002, while the LVIS subset uses seed 20261003.
Both are therefore disjoint from the tuning data. The two ablation subsets
share 40 image identifiers because LVIS-minival itself contains images from
COCO val2017; nevertheless, they are independently sampled and are evaluated
against the annotations of their respective benchmarks.

\subsection{Vocabulary Caching}

For each image, MLLM, and prompt combination, the generated object vocabulary
is cached before detector evaluation. The grounding models then consume this
cached vocabulary directly rather than independently querying the MLLM.

Consequently, all open-vocabulary detectors paired with a given MLLM receive
exactly the same generated label list for the same image and prompt. This
isolates the effect of the grounding model in detector comparisons and prevents
variation in MLLM generation from confounding differences between detectors.
It also allows the generated vocabulary to be reused when evaluating multiple
grounding models.

\subsection{Vocabulary Composition}

All experiments reported in this paper use only the vocabulary generated by
the MLLM. No user-specified category set is merged with the generated
vocabulary, and no embedding-based semantic filtering or category-merging
procedure is applied before grounding. The generated class names are therefore
passed directly to the open-vocabulary detector without additional
modification.

Thus, the reported evaluation directly measures the two-component formulation
described in the main paper: the MLLM determines the image-specific concepts
to detect, and the open-vocabulary detector determines where those concepts
occur.

\end{document}